\documentclass{article}

\usepackage{amssymb}

\usepackage{graphicx}

\usepackage{float}        
\usepackage{subfigure}
\title{Making Early Predictions of the Accuracy of Machine Learning Applications}

\author{ J.E. Smith and P. Caleb-Solly,\\
Department of Computer Science and Creative Technologies,\\
University of the West of England, 
Bristol, BS16 1QY, UK \\ 
 james.smith@uwe.ac.uk
\and
 M.A. Tahir\\
 Northumbria University, UK
\and
D. Sannen\\
Flanders Mechatronic Technology Centre,\\
 Leuven, Belgium
\and
H. Van Brussel\\
Department of Mechanical Engineering,\\
Katholieke Universiteit Leuven,\\
B-3001 Heverlee, Belgium
}

\begin{document}

\maketitle

\begin{abstract}

The  accuracy  of machine learning systems is a widely studied research topic. Established techniques such as cross-validation predict the accuracy on unseen data of the classifier produced by applying a given learning method to a given training data set. However, they do not predict whether incurring the cost of obtaining more data and undergoing further training will lead to higher accuracy. In this paper we investigate techniques for making such early predictions. We note that when a machine learning algorithm is presented with a training set the classifier produced, and hence its error, will depend on the characteristics of the algorithm, on training set's size, and also on its specific composition. In particular we hypothesise that if a number of classifiers are produced, and their observed error is decomposed into bias and variance terms, then although these components may behave differently, their behaviour may be predictable.

We test our hypothesis by building models that, given a measurement  taken from the classifier created from a limited number of samples, predict the values that would be measured from the classifier produced when the full data set is presented. We create separate models for bias, variance and total error. Our models are built from the results of applying ten different machine learning algorithms to a range of data sets, and tested with ``unseen'' algorithms and datasets. We analyse the results for various numbers of initial training samples, and total dataset sizes.  Results show that our predictions are very highly correlated with the values observed after undertaking the extra training. Finally we consider the more complex case where an ensemble of heterogeneous classifiers is trained, and show how we can accurately estimate an upper bound on the accuracy achievable after further training.

\end{abstract}





%
\section{Introduction}\label{intro}

Predicting the accuracy of a trained machine learning system when presented with previously unseen test data is a widely studied research topic. Techniques such as cross-validation are well established and understood both theoretically and empirically e.g.\ \cite{HastieTibshiraniFriedman01,Stone74}. However, these techniques predict the accuracy on unseen data \emph{given the existing training set}. For example  $N$-fold Cross Validation ($NCV$) averages the fitness estimated from $N$ runs, each  using a proportion $1-1/N$ of the available data to train a classifier and $1/N$ to evaluate it. Therefore repeating with different values of $N$ can give the user some indication of how the error rate changed as the training set increased to the current size, since lower values of $N$ effectively equate to smaller training sets. However, $NCV$ does not predict  what  accuracy might be achievable after further training. Thus if the current accuracy is not acceptable, and obtaining data comes at cost, $NCV$ and similar techniques do not offer any insights into whether it is worth incurring the cost of further training.

This is of more than theoretical interest, because the successful application of machine learning techniques to ``real-world'' problems places various demands on the collaborators. Not only must the management of the industrial or commercial partner must be sufficiently convinced  of the potential benefits that they are prepared to invest money in equipment and time, but  vitally, there must be a significant investment in time and commitment from the end-users in order to provide training data from which the system can  learn. This poses a problem if the system developed is not sufficiently accurate, as the users and management may view their input as wasted effort, and lose faith with the process.

In some cases this effort may be re-usable -- for example, the user has been labelling training examples that can be stored in their original form, and which come from a fairly stationary distribution. However, this is frequently not the case. For example, in many applications it may not be practical to store the physical training examples, rather it  is necessary to characterise them by a number of variables. If the failure of the Machine Learning system in such cases stems from an inappropriate or inadequate choice of descriptors, then the whole process must be repeated. Not only has the user's input been a costly waste of time and effort, but there may be a loss of faith in the process which can manifest in reduced attention and consistency when classifying further samples. To give a concrete example from the field of diagnostic visual inspection (e.g.\ manufacturing process control  or medical images), it frequently turns out that it is not sufficient to store each relevant image -- other information is necessary such as process variables, or patients' history. If this data is not captured at the same time, and is not recoverable post-hoc,  then the effort of collecting and labelling the database of examples has been wasted.

A significant factor that would help in gaining confidence and trust from  end-users would be the ability to quickly and accurately predict whether the learning process was going to be successful. Perhaps more importantly from a commercial viewpoint, it would be extremely valuable to have an early warning that the users can save their effort while the system designer refines the choice of data, algorithms etc.

In this paper we investigate a technique for making such early predictions of future error rates. We will consider that we are given $n$ samples, and that the system is still learning and refining its model at this stage. We are interested in predicting what final accuracy might be achievable if the users were to invest the time to create $n'$ more samples. This leads us to focus on two questions. First, what are the most appropriate  descriptors of the system's behaviour after some limited number $n$ of samples, and then later after an additional $n'$ samples? Second, is it possible to find useful relationships for predicting the second of these quantities from the first?

Theoretical studies, backed up by empirical results, have suggested that the total error rate follows a power-law relationship, diminishing as extra training samples are provided. While these theoretic bounds on error are rather loose,  they provide motivation for investigating practical approaches for quickly and reliably estimating the error rate that may be observed after future training.
In general the error will be a complicated function, but the hypothesis of this paper is that we can deal with it more easily if we decompose it into a number of more stable functions. Therefore this paper concentrates on the use of the well-known bias-variance decomposition \cite{Kohavi,Brian} as a source of predictors when an algorithm is used to build a classification model from a dataset.
Specifically, our hypothesis is that if the observed error is decomposed into bias and variance terms, then although these components may behave differently, their behaviour may be individually predictable.

To test our hypothesis we first apply a range of algorithms to  a variety of datasets, for each combination periodically estimating the error components as more training samples are introduced, until the full dataset has been used. All of the data arising from this (rather lengthy) process is merged and  regression analysis techniques are applied to produce three sets of predictive models - one each for bias, variance and total error.  Each of these  models takes as input a measurement obtained from the classifier produced when only a few samples ($n$) from a dataset have been presented to the learning algorithm, and predicts the value after all  samples have been applied ($n+n'$).
As the data has been merged, the intention is that these models are algorithm-dataset independent.  We examine the stability and valid range of these models, and evaluate their predictive power when using previously unseen datasets and algorithms. Moving on to consider trainable ensembles of different classifiers, we show how a similar approach can be applied to obtain estimates on the upper bound of the achievable accuracy, which can predict the progression of the ensemble's performance.

The rest of this paper proceeds as follows. In Section~\ref{sec:review} we review related work in the field, including the bias-variance decomposition of error that we will use.   Following that, Section~\ref{methodology} describes the experimental methodology used to collect the initial statistics, and test the resulting models.  Sections~\ref{single_results} and~\ref{unseen} describe and discuss the results obtained. In Section~\ref{ensembles} we show how this approach may be extended to predict the future accuracy of trainable ensembles of classifiers. Finally in Section~\ref{conclusions} we draw some conclusions and suggestions for further work.

\section{Background}\label{sec:review}
\subsection{Notation}\label{subsec:notation}
For the sake of clarity we will use a standard notation throughout this paper,  reinterpreting results from other authors as necessary.

We assume classification tasks, where we are given an instance space $X$ and a predicted categorical variable $Y$. The ``true'' underlying function $F$ is a mapping $F:X \rightarrow Y$.

Let $D$ be the set of all possible training sets of size $n$ sampled from the instance space $X$,
and $d \in D= \{(x_1,y_1),(x_2,y_2), \ldots ,(x_n, y_n)\}$.

When a machine learning algorithm $C$ is presented with $d$ it creates a classifier, which we may view as a hypothesis about the underlying mapping: $H_{Cd}: X \rightarrow Y$. The subscripts  $C$ and $d$ make it explicit that the specific classifier $H$ induced depends on the learning algorithm and the training set. For a specific learning algorithm $C$ the set of classifiers that it can induce is denoted $\mathcal{H}$.

We consider a $0/1$ misclassification error - in other words the error is zero if $H$ correctly predicts the true class of an item $x \in X$, and 1 otherwise. More formally the misclassification cost of a single data item $x$ with a specific classifier $H$ is:
\begin{equation} \label{eq:cost}
Cost (H_{Cd},x) = \left \{ \begin{array}{llll}  0 & H_{Cd}(x) & =& F(x) \\		1 & H_{Cd}(x) & \neq& F(x) \\ \end{array}  \right .
\end{equation}
The expected error of the classifier created from  $n$ data points is then given by integrating over $X$ and $d$, taking into account their conditional likelihood, i.e.:
\begin{equation} \label{eq:error-general}
Error(H_{Cn},X)  = \int_{x \in X, d \in D} P(x) P(d|n)Cost(H_{Cd},x)
\end{equation}

In practice of course it is not possible to exactly measure the true error,  so approaches such as bootstrapping, hold-out and cross-validation are used to estimate the error, given a finite sized set of examples. We will use the lower case $error$ to denote an estimation is being used for the true error.

\subsection{Relationship to Other Work}\label{previous}

Cortes \textit{et al} \cite{cortes94} presented an empirical study where they characterised the behaviour of classification algorithms using ``learning curves''. These suggest that the predicted error of the classifier after $n$ samples have been presented will follow a power-law distribution in $n$:
\begin{equation} \label{eq:cortes-pwerlaw}
error(n) = an^{-\alpha} +b
\end{equation}
\noindent where the constants $a$ (the learning rate), $\alpha$ (the decay rate) and $b$ (the asymptotic Bayes error rate) depend on the particular combination of classification algorithm and data set, but $\alpha$ is usually close to, or less than one.
This suggests that given a particular classifier-dataset combination, it should be possible to commence training, take periodic estimates of the error as $n$ increased, and then use regression to find values for $a,b,\alpha$ that fit the data, and can be used for predicted future error rates. Mukerhjee \textit{et al} \cite{mukherjee03} have pointed out a problem with this approach, namely that for low values of $n$ the estimated error rates are subject to high variability, which leads to significant deviations when fitting the power-law curve. They have presented an extension of the method which uses a ``significance permutation test'' to establish the significance of the observed classifier error prior to curve fitting.

These results fit in with theoretical bounds from  ``Probably Approximately Correct'' (PAC) theory such as those presented by Vapnik in \cite{vapnikbook}.
These begin with  the assumption that a training set $d = \{x_k,y_k\}, 1 \leq k \leq n, y_k \in \{-1,1\}$ is drawn independently and identically distributed (iid) from a data set, and that future training and test data will be drawn from the dataset in the same way.
Given the restriction $Y = \{-1,1\}$, the test error $Error(H_{Cn})$, (the probability of misclassification) is defined to be:
\begin{equation}
Error(H_{Cn}) = E\left [ \frac{1}{2}\mid F(x) - H_{Cn}(x) \mid \right ]
\end{equation}
where the division by two maps differences in $Y$ onto costs, and the $H_{Cn}$ denotes that we are taking the expectation for the general case. The current empirically measured training error $error(H_{Cd})$ is:
\begin{equation}
error(H_{Cd}) = \frac{1}{n} \sum^n_{k=1} \frac{1}{2} \mid y_k - H_{Cd}(x_k)\mid
\end{equation}

If  $\psi$ represents the VC-dimension of classification algorithm $C$, and $0 \leq \eta \leq 1$  Vapnik \cite{vapnikbook} showed that with probability $1 - \eta$:
\begin{equation}\label{eq:vc-dim-error}
Error(H_{Cn}) \leq error(H_{Cd}) + \sqrt{ \frac{\psi log(2n)  +\psi (1- log{\psi}) - log(\eta/4)}{n}}
\end{equation}

Effectively this equation makes explicit an assumption that machine learning algorithms inherently produce classifiers which overfit the available training data. The VC-dimension $\psi$ may be thought of as the ``power'' of the machine learning algorithm $C$ -- it is the maximum number of points that can  be arranged so that $C$ can ``shatter'' them. Equation~\ref{eq:vc-dim-error} makes it clear  that more powerful algorithms (higher $\psi$) are more likely to overfit the data, and so it may be used, for example, as grounds to select between two algorithms which produce the same training error but have different complexity (related to  $\psi$).
It also makes explicit the dependency on $n$: for a given training set error, the maximum amount by which this will underestimate the true error decreases by approximately $\sqrt{ \psi log{n} / n}$.

However, in practice these bounds tend to be rather ``loose''. There have been other more recent developments in Statistical Learning Theory which use a similar approach but exploit Rademacher complexity to provide tighter bounds,  such as those   
in \cite{BartlettMendelson02, BartlettBoucheronLugosi2002, Bousquet2004, Boucheron2005}. Common to all of these approaches,as with the use of VC-dimension results,  is the idea that on the basis of the available training data,  an algorithm selects a classifier $H_{Cd}$ from some class $\mathcal{H}$ available to it.  To analyse the learning outcomes,  the "error" observed when the training data is classified by $H_{Cd}$ is broken down into the Bayes optimal error (which cannot be avoided)  plus an amount by which best ($H* \in \mathcal{H}$) in the current class of classifiers would be more than Bayes optimal,  plus an amount by which the classifier $H_{Cd}$ currently estimated by the algorithm to be "best" is different to the actual best $H*$.  Thus for example, approaches such as Structural Risk Minimisation can be thought of as principled methods for increasing the size/complexity of the current class of classifiers $\mathcal{H}$ until it includes the Bayes optimal classifier. 

The  underlying assumption is that the error is estimated using the current training set,  and that this almost certainly overfits the true underlying distribution (i.e. $H_{Cd} \neq H*$)  so  
the current estimates of error for the chosen classifier $H_{Cd}$ will be less than the "true" error that would be seen if it was applied to the whole data distribution.
Therefore bounds are derived which describe the extent to which the error on the training set underestimates the true error.  Since this can be described in terms of the search problem of identifying $H* \in \mathcal{H}$,  it is understandable that they take into account the amount of information available to the search algorithm - i.e. the size $n$ of the training set. 

While this is a valid and worthwhile line of theoretical research,  we would argue that it is not currently as useful for the practioner.  Consider the example of a user who is highly skilled in their domain,  but knows nothing about  Machine Learning,  and is providing the training examples from which a classifier is constructed. The theory above effectively says: \textit{ "Based on what you have told me,  I've built a classifier which seems to have an error rate of $x$\%. I can tell you with what probability the "true" error rate is  worse than $x+y$\%,  for any positive $y$"}.    If they have provided enough labelled data items to create what appears to be an accurate classifier, then this is valuable.  However,  if they are still early on in the process,  and the current error rates are high,  it gives no clues as to whether they will drop.  Instead we attempt to provide heuristics that answer a different question: \textit{"Based on what you have told me,  I've built some  classifiers and although the current error rate is $x$\% it will probably drop to $y$\%  where $y \leq x$."}

To do this, we note that the analysis above  relates the true test error to a specific estimated error  from a given training set size, and as discussed above the variance in the predicted error depends strongly on $n$. This has prompted us to examine different formulations that explicitly decompose the error into terms arising from the inherent bias of the algorithm (related to its VC-dimension, or to the difference between $H*$ and the Bayes optimal classifier)) and the variability arising from the choice of $d \in D$.

\subsection{Bias-Variance Decomposition}\label{bv_decomposition}
 A number of recent studies have shown that the decomposition of a classifier's error into bias and variance terms can provide considerable insight into the prediction of the performance of the classifier \cite{Kohavi,Brian}. Originally, it was proposed for regression \cite{Geman} but later, this decomposition has been successfully adapted for classification \cite{Kohavi,Brian,Rodriguuez}. While a single definition of bias and variance is adopted for regression, there is considerable debate about how the definition can be extended to classification \cite{Kohavi,Breiman,Domingos,Friedman,James,Kong}. In this paper, we use Kohavi and Wolpert's \cite{Kohavi} definition of bias and variance on the basis that it is the most widely used definition \cite{Webb1,Webb}, and has strictly non-negative variance terms.

 Kohavi and Wolpert define bias, variance and noise as follows \cite{Kohavi}:

\begin{description}
  \item[Squared Bias:] ``This quantity measures how closely the learning algorithm's average guess (over all possible training sets of the given training set size) matches the target''.
  \item[Variance:] ``This quantity measures how much the learning algorithm's guess bounces around for the different training sets of the given size''.
  \item[Intrinsic noise:] ``This quantity is a lower bound on the expected cost of any learning algorithm. It is the expected cost of the Bayes-optimal classifier''.
\end{description}

Given these definitions, we can restate Eq.~\ref{eq:error-general} as:

\begin{equation}\label{kohaviandwolpert}
    Error(H_{C,n}) = \int_{x \in X}P(x)(\sigma_{x}^{2}+bias_{x}^{2} + variance_{x})
\end{equation}
Assuming a fixed cardinality for $Y$ (finite set of classes), and noting $D$ has finite cardinality,  the summation terms are:\\
\begin{eqnarray}
 Bias_{x}^{2} &=& \frac{1}{2}\sum_{d \in D}P(d|F,n)\sum_{y \in Y}\left[P(F(x) = y)-P(H_{Cd}(x) =y)\right]^2, \nonumber \\
 Variance_{x} &=& \frac{1}{2}-\frac{1}{2}\sum_{y \in Y} \sum_{d \in D}P(d|F,n)P(H_{Cd}(x)=y)^2, \nonumber \\
\sigma_{x}^{2} &=& \frac{1}{2}-\frac{1}{2}\sum_{y \in Y} P(F(x)=y)^2. \nonumber
\end{eqnarray}
\noindent where the terms  $P(F(x)=y), P(H_{Cd}(x)=y), P(d|F,n)$ make explicit that some terms are conditional probability distributions since the Bayes error may be non-zero, the classification output may not be crisp, and the specific choice of training set depends on the underlying function and the number of samples.

\subsection{Bias as an Upper Limit on Accuracy}
An alternative perspective on the above analysis is that the bias term reflects an inherent limit on a classifier's accuracy resulting from the way in which it forms decision boundaries. For example,  an elliptical class boundary can never be exactly replicated by a classifier which divides the space using axis-parallel decisions. A number of studies have been made confirming the intuitive idea that the size of variance term drops as the number of training samples increases, whereas the estimated bias remains more stable, e.g.\ \cite{Brian}. Therefore we can treat the sum of the inherent noise and bias terms as an upper limit on the achievable accuracy for a given classifier. Noting that in many prior works it is assumed that the inherent noise term is zero, and that for a single classifier it is not possible to distinguish between inherent noise and bias, we hereafter adopt the convention of referring to these collectively as bias.

\subsection{Hypothesis of the Paper}
The hypothesis of the main part of this paper is that values of the bias and variance components estimated  after $n$ training samples can be used to provide accurate predictions for their values after $n+n'$ samples, and hence for the final error rate observed. To do this prediction we use statistical models built from a range of dataset-algorithm combinations. To create the model data we repeatedly draw training and test sets from the $n$ samples from which we can estimate the total error, together with its bias and variance components. This raises the issue of how we should do this repeated process.

\subsection{Prediction Methodology and Sampling Considerations}\label{sampling}
If the variables in $X$ are continuous, or unbounded integers, then the underlying distribution over which the classifier may have to generalise is of course infinite.  For bounded integer or categorical variables, the number of potential training sets of size $n$ drawn iid from an underlying distribution of $X$ is of size $|D| = |X|!/n! (|X|-n)!$, so in practice even for non-trivial datasets it is not possible to evaluate all possible training sets $d$ of size $n$. However
the success (or otherwise) of the approach proposed in this paper  depends on the accuracy with which we can predict error components, particularly for when the training set sizes are low. This immediately raises the question of finding the most appropriate methodology for estimating the values of those quantities.
To give a simple example of why this is important, a later result in this paper partially relies on being able to distinguish between those data items that are \emph{always} going to be misclassified by a given classifier, and those which will \emph{sometimes} be misclassified, depending on the choice of training set. Since the well known $N$-fold cross-validation approach  only classifies each data item once, it does not permit this type of decomposition and cannot be used. In a preliminary paper \cite{smithtahir07} we have examined two possible approaches: the ``hold-out'' method proposed by Kohavi and Wolpert \cite{Kohavi} and the ``Sub-Samples Cross Validation'' (SSCV) method proposed by  Webb and Conilione \cite{Webb}. The latter have argued that the hold-out approach proposed in \cite{Kohavi} is fundamentally flawed, partly because it results in small training sets, leading to instability in the estimates it derives. This was confirmed by our results  \cite{smithtahir07} which  showed that the stability of the estimates, and hence the accuracy of the resulting prediction was far higher for the Sub-Sampling method. Therefore  we restrict ourselves to this approach.

The SSCV procedure is designed to address weaknesses in to both the hold-out and bootstrap procedures by providing a greater degree of variability between training sets. In essence, this procedure repeats $N$-fold CV $l$ times, thus ensuring that each sample $x$ from the training set of size $n$ is classified $l$ times by the classifier $i$. The true $Bias_{ix}$ and $Variance_{ix}$ can be estimated as $bias_{ixn}$ and $variance_{ixn}$ from the resulting set of classifications. The final bias and variance is estimated from the average of all $x \in D$ \cite{Webb1,Webb}, thus using all $n'$ samples.

\section{Experimental Methodology}\label{methodology}
The following sections describe our choice of experimental methodology, algorithms and data sets. Please note the distinction between those datasets and algorithms used to provide the data from which the statistical models were built, and those which were only used for evaluation purposes.

\subsection{Choice of Classifiers:}\label{methodology:choice}
In order to obtain the data for modelling ten different classification algorithms were selected, each with different bias and variance characteristics. These were: Naive Bayes (NAI) \cite{Duda}, C4.5 (C4.5) \cite{Quinlan93}, Nearest Neighbour (1NN) \cite{Cover}, Bagging (BAG) \cite{Leo1}, AdaBoost (ADA) \cite{Yoav}, Random Forest (RAF) \cite{Leo}, Decision Table (DTB) \cite{Kohavi1}, Bayes Network (BAN) \cite{Duda}, Support Vector Learning (SMO) \cite{Platt}, and Ripple-Down Rule learner (RID) \cite{Weka}. Note that this set includes two methods for creating ensembles: AdaBoost (using Decision Stumps as the base classifier) and Bagging (using a decision tree with reduced error pruning). In these cases, since we are solely interested in the outputs, we treat the ensemble as a single entity, rather than attempt a bias-variance-noise-covariance decomposition.

In the evaluation we investigate how well the models can extrapolate when new classifiers are used. Five classifiers, again with different bias-variance trade-offs, are used for this analysis namely CART (CART) \cite{LBreiman}, RandomSubSpace (RSS) \cite{Tin}, Logistic (LOG) \cite{LeCessie}, KNN with $K=3$ (3NN), and Complement Naive Bayes (CNB) \cite{Rennie}.

All these classifiers are implemented in the WEKA library \cite{Weka}, and the default parameters in WEKA are used for each classifier.

\subsection{Data Sets}

The data collection required to build the statistical models is carried out on data sets derived from four Artificial and five real-world visual surface inspection problems from the DynaVis project\footnote{\texttt{www.dynavis.org}}. Each artificial problem consists of 13000 contrast images created by a tuneable randomised image generator. Class labels (good/bad) were assigned to the images by using different sets of rules of increasing complexity acting on the generator. The real world data sets came from CD-imprint and egg inspection problems. There are 1534 CD images, each  labelled by 4 different operators, and 4238 labelled images from the egg inspection problem. The same set of image processing routines are applied to segment and measure regions of interest (ROI) in each image. From each set of images are derived 2 data sets. The first has 17 features describing global characteristics of the image and the ROI it contains. In the second these are augmented by the maximum value (over all the ROI) for each of 57 ROI descriptors. Adding the labels available provides a total of 18 different data sets with a range of dimensionality and cardinality.

To build the models we used 14 of the data sets: the six derived from the first three artificial image sets, the six from the CD images labeled by the first three operators and the two from the egg data. The remaining four data sets, derived from the fourth artificial image set, and the CD labelled by Operator 4 are reserved for evaluation purposes, as are three example datasets selected from the UCI repository \cite{Repository}.

In each case we took $n' = total\_set\_size  - 1000$, so $n'$ differs between data sets. 
\subsection{Procedure for Building Models}

\begin{figure}[tbp]
\setlength{\unitlength}{0.47cm} 
\centering 
\begin{picture}(30,15) 

\put(2,13){$n$ samples}
\put(2,12){SSCV}
\put(2.5,10){\vector(2,1){5}}

\put(7.5, 10){\framebox(10,5){\begin{tabular}{l}Estimated Bias ($b_{ixn}$) \\ Estimated Variance ($v_{ixn}$)\\ Estimated Error ($e_{ixn}$)\end{tabular}}}

\put(17.5,12.5){\vector(4,-1){5}}
\put(0,5){\framebox(5,5){\begin{tabular}{l}	Classifier $i$ \\ Dataset $x$ \end{tabular}}}
\put(18,3){\framebox(12,8){\begin{tabular}{c}	Pair observations and \\
									create Regression Models\\
									 $b_{ixn'} = a_{Bn}* b_{ixn} + b_{Bn}$ \\
									 $v_{ixn'} = a_{Vn}* v_{ixn} + b_{Vn}$\\
									 $e_{ixn'} = a_{En}* e_{ixn} + b_{En}$ \end{tabular}}}

\put(1,2){$n + n'$ samples}
\put(1,1){ SSCV}
\put(2.5,5){\vector(2,-1){5}}
\put(7.5, 0){\framebox(10,5){\begin{tabular}{l}Estimated Bias ($b_{ixn'}$) \\ Estimated Variance ($v_{ixn'}$)\\ Estimated Error ($e_{ixn'}$)\end{tabular}}}
\put(17.5,1.5){\vector(4,1){5}}
\end{picture}
   \caption{Methodology for creating predictive models. This is repeated for $n \in\{100,200, \ldots, 1000\}$\label{fig:regression}}
\end{figure}
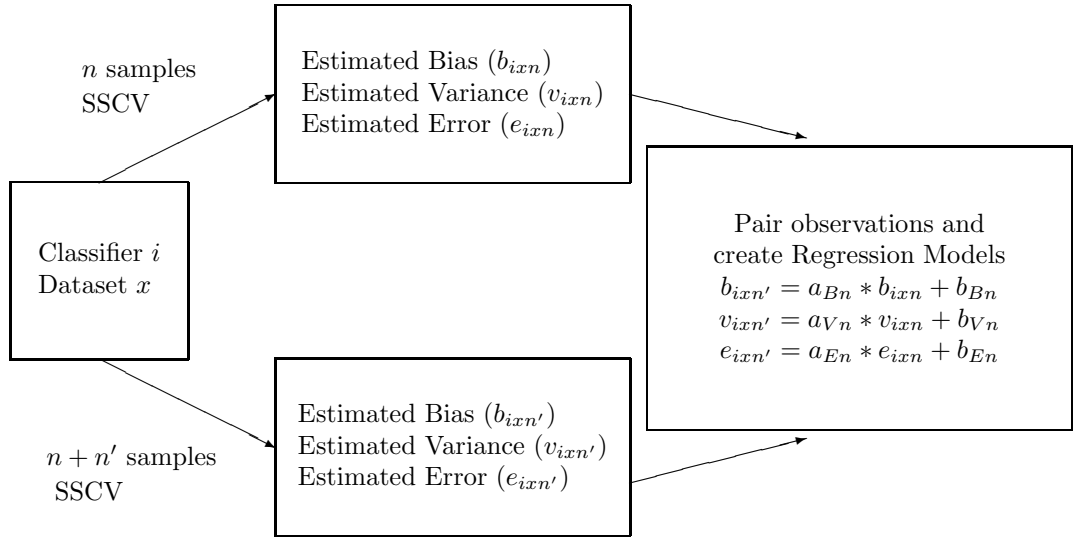

Our experimental procedure is as follows:
\begin{itemize}
\item For each dataset we used SSCV to estimate the values of error, bias and components using the first $n \in \{100,200,\ldots,1000\}$ samples.
\item For each dataset we then  used SSCV to estimate the values of error, bias and variance using all of the samples in the dataset. Note that this results in different values of $n'$ for different datasets.  Note also that we do not use a separate ``test set''. We consider that since one is always making estimates of the error on unseen data it is more consistent to relate estimates of the error at different points in training \emph{using the same estimation methodology}.
\item All of the collected data was pooled, we applied linear regression to create models of the form $V_{(n+n')} = a \cdot V_n +b$ where $V \in \{total\; error, bias, variance\}$. In these models $V_n$ is the independent variable, $V_{(n + n')}$ the dependent variable and  $a$ and $b$ are the constants to be estimated by the linear regression procedure. We compute the coefficient of determination $R^2$ to measure how well the simple linear model explains the variability of the independent variable, and hence the quality of the resulting predictions -- the closer $R^2$ is to 1, the better is the prediction.
\item Note that  this gives us two ways of predicting the error -- either directly or by summing the predictions for bias and variance.
\end{itemize}

Figure~\ref{fig:regression} shows this process for a single value of $n$.

We would like to re-iterate for the sake of clarity that we are not building models which relate error, bias and variance as a function of the number of training samples $n$.  In that case it would certainly be true that by the two models (bias as a function of $n$) and (variance as a function of $n$) could be combined into a single linear model (error as a function of $n$).  As the welath of theoretical work described above shows,  there is ample evidence to sugest that no simple predictive lnear model exists.  
  Instead we are building and combining linear models of the form \textit{ future bias/variance/error as a function of current bias/variance/error}  and seeing how the predictive power of these models changes as a  result of the value of   $n$.

These linear models are of course an extremely simple way of modelling the relationship between our various predictors, and more sophisticated techniques exist in the fields of statistics and also Machine Learning. However, as the results will show it is sufficient for our purposes.  An obvious candidate for future work is to consider approaches which will give us confidence intervals on the predicted error, as this will fit in better with the concept of providing an upper bound on the achievable accuracy.

\section{Results: Explanatory Power of Models}\label{single_results}

\begin{figure}[tbp]
    \centering
\includegraphics*[height=0.3\textheight]{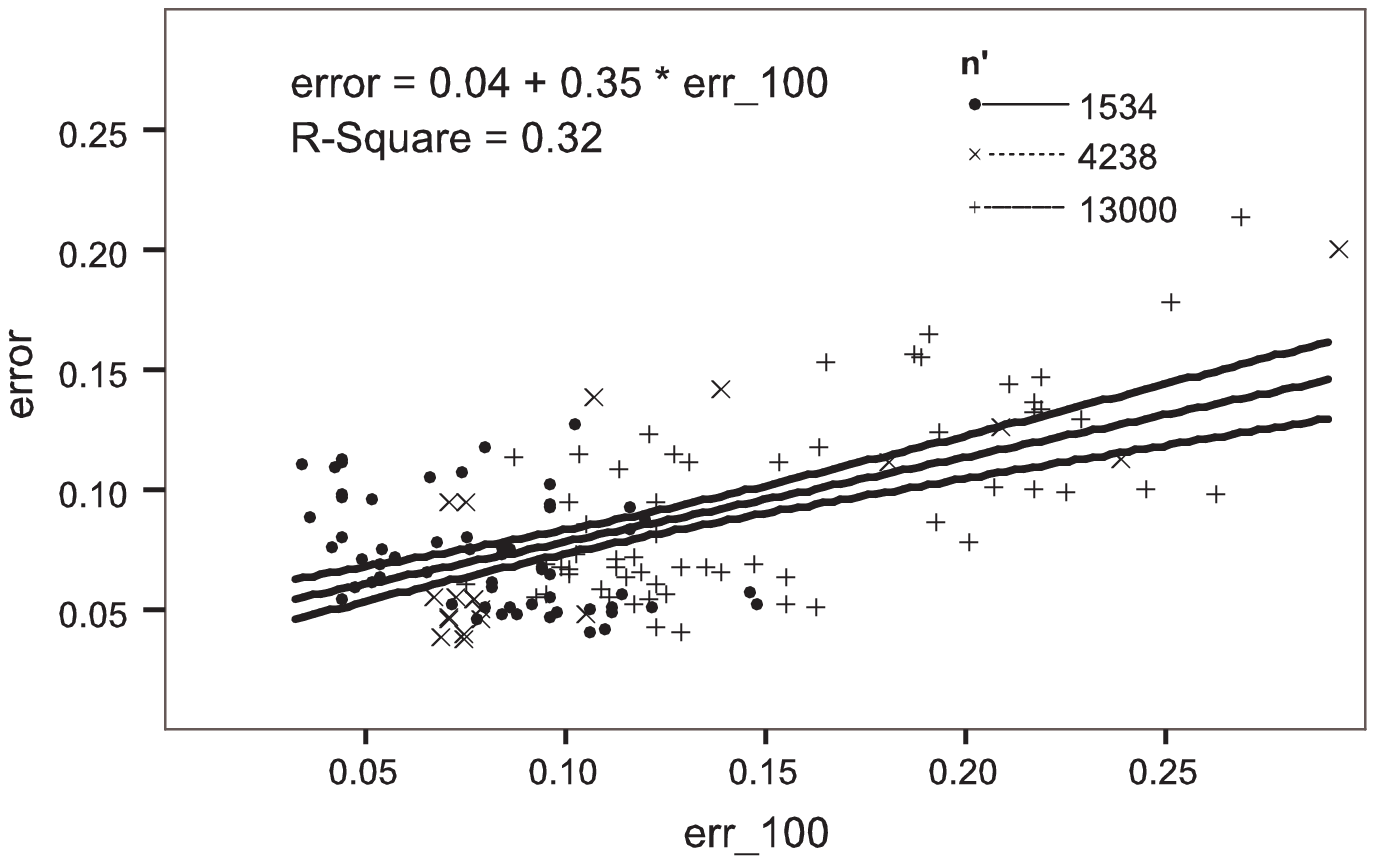}
\includegraphics*[height=0.3\textheight]{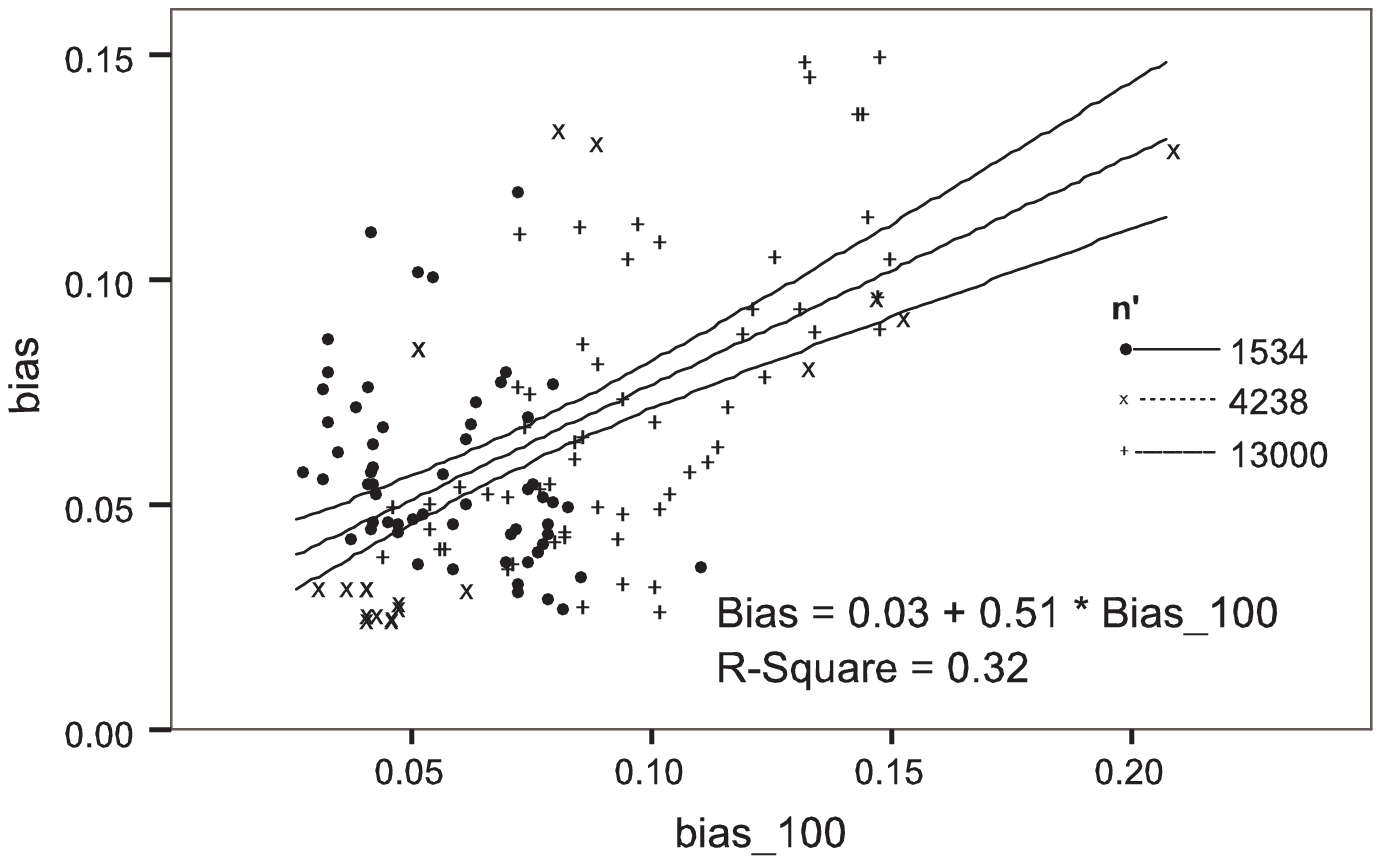}
\includegraphics*[height=0.3\textheight]{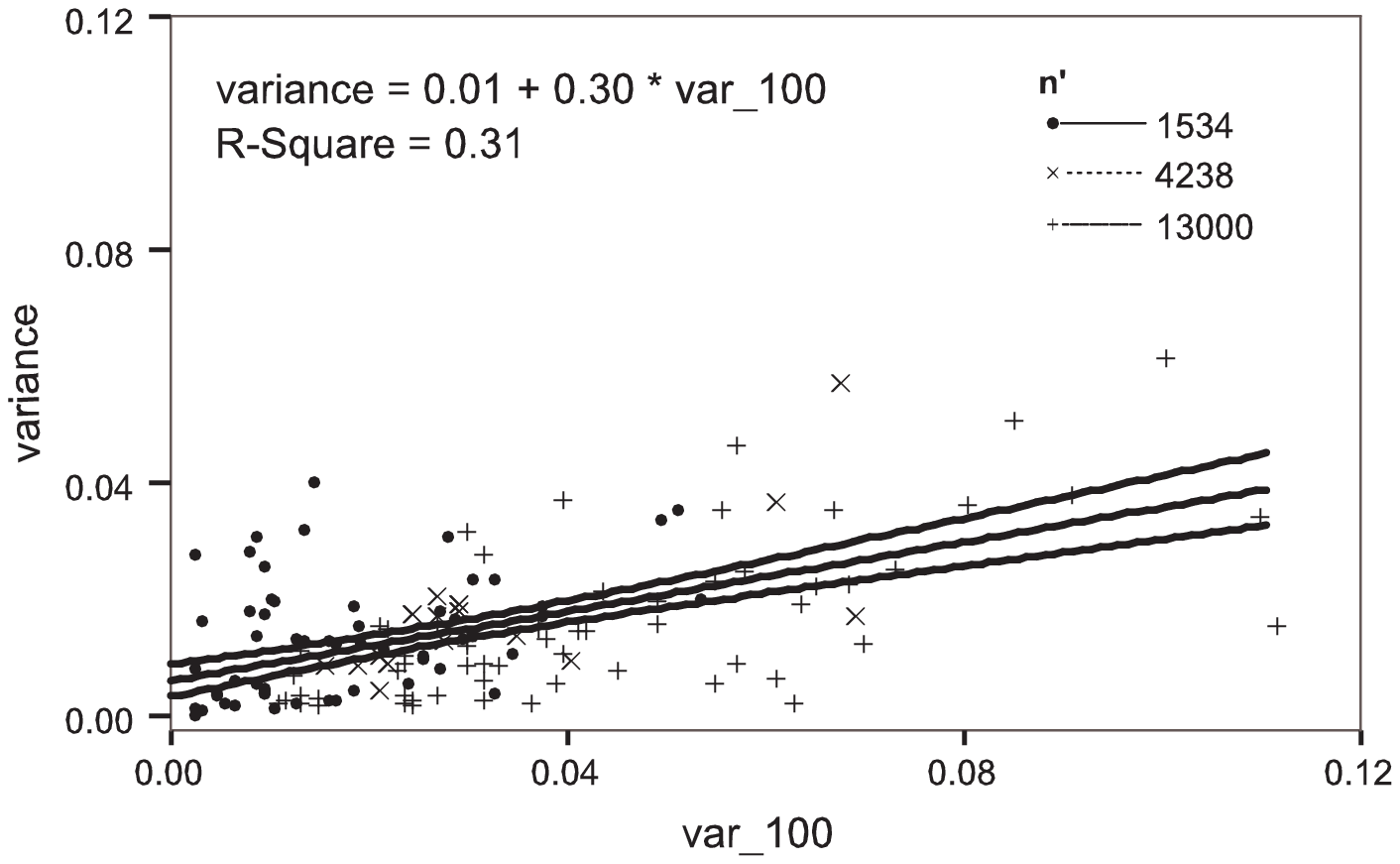}
   \caption{Scatter plots of the Error (top), Bias (middle) and Variance (bottom) estimated after 100  samples  ($x$-axis) and the same descriptors estimated using all samples ($y$-axis), together with results from linear regression. Absolute values on individual plots vary, but in each case the $x$ and $y$ axes scale over the same range (so a 1:1 correspondence would form a diagonal of the plot). \label{figure-real-webb100}}
\end{figure}

\begin{figure}[tbp]
    \centering
\includegraphics*[height=0.3\textheight]{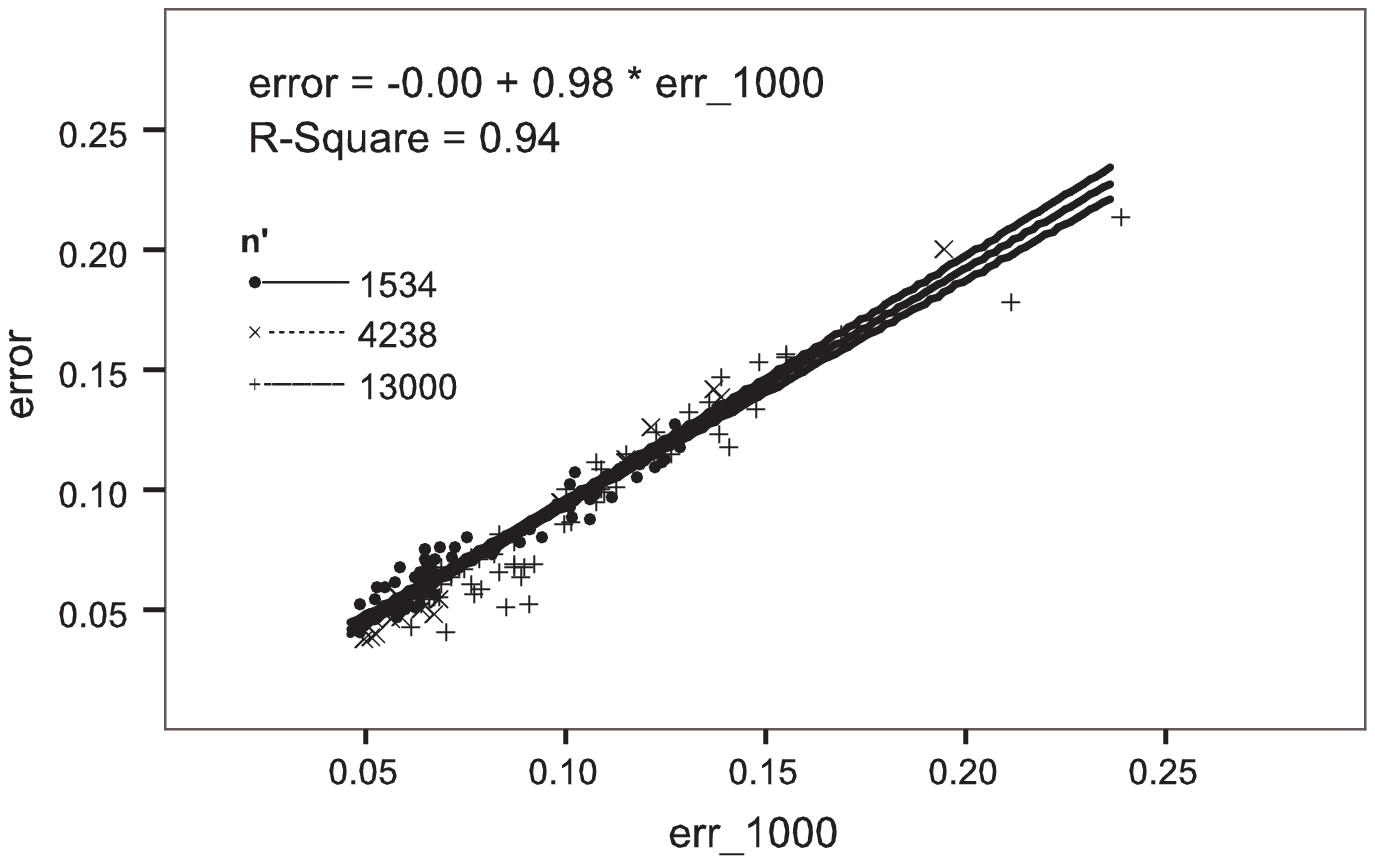} \\
\includegraphics*[height=0.3\textheight]{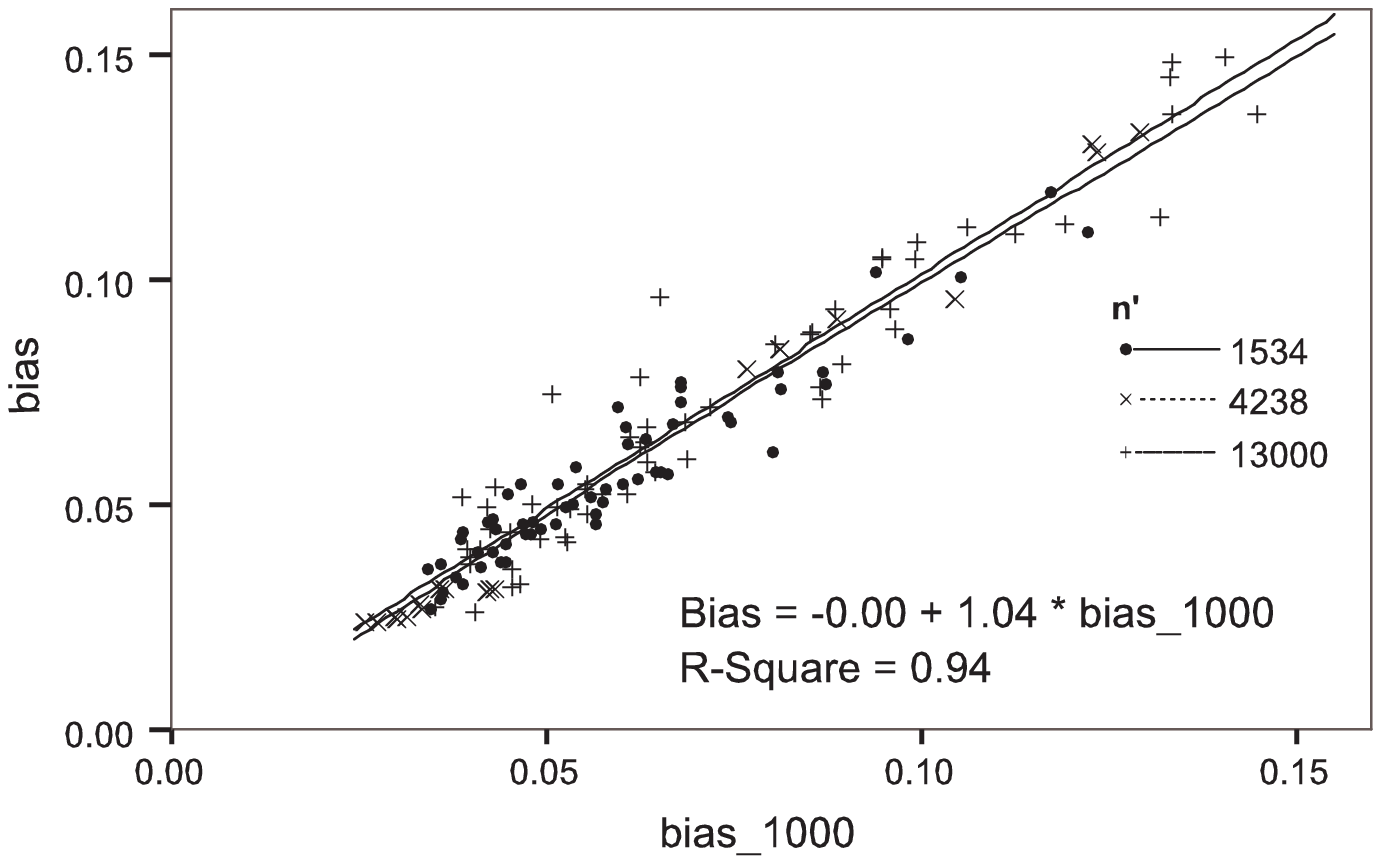} \\
\includegraphics*[height=0.3\textheight]{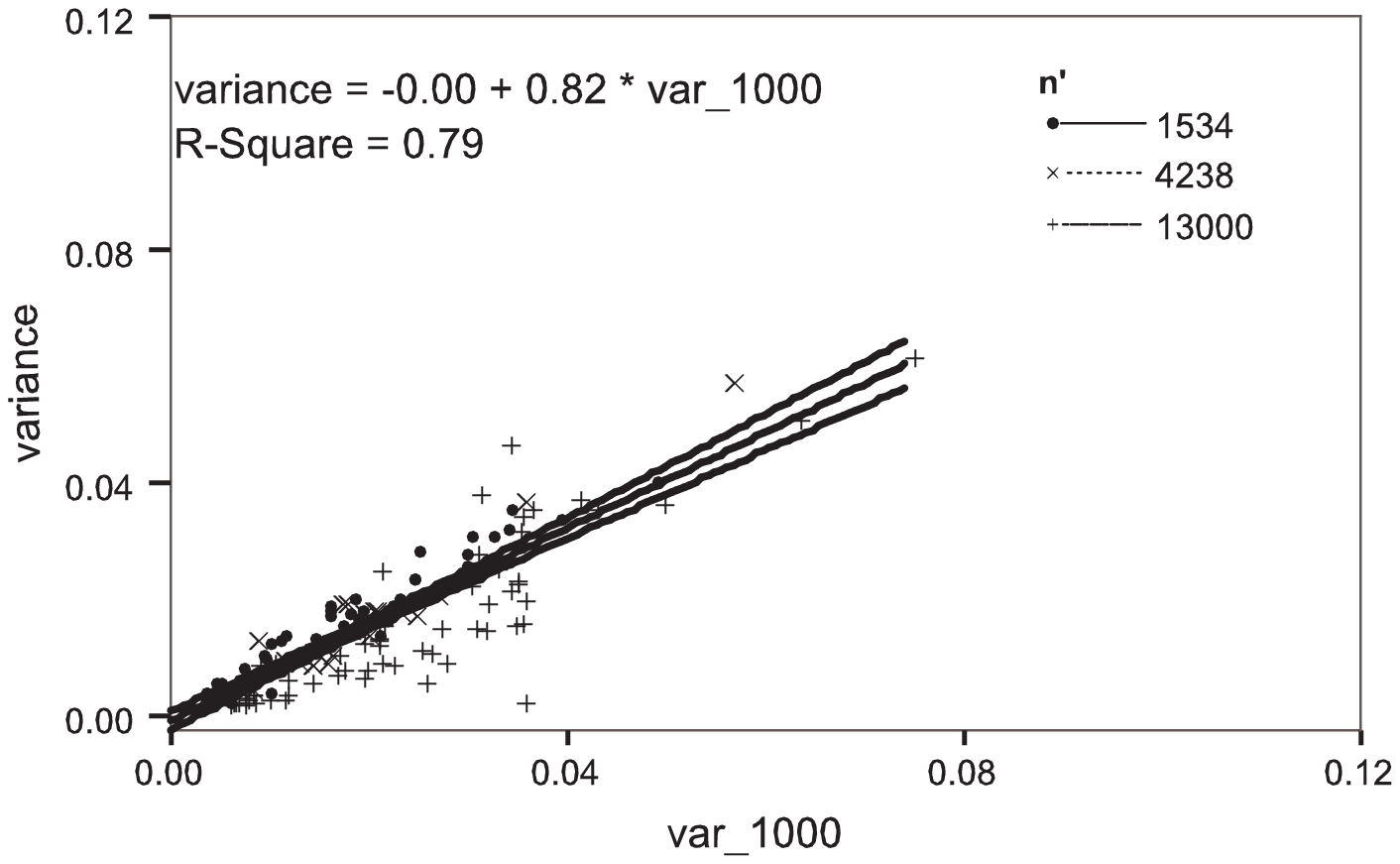} \\
\caption{Scatter plots of the Error (top), Bias (middle) and Variance (bottom) estimated after 1000 (right) samples  ($x$-axis) and the same descriptors estimated using all samples ($y$-axis), together with results from linear regression. Absolute values on individual plots vary, but in each case the $x$ and $y$ axes scale over the same range (so a 1:1 correspondence would form a diagonal of the plot). \label{figure-real-webb1000}}
\end{figure}

Figures~\ref{figure-real-webb100} and \ref{figure-real-webb1000} show scatter plots of the values for error, bias and variance as measured after $n \in \{100,1000\}$ samples and after all samples. Different markers indicate different total numbers of samples. Note that in each case the same range is used for $x$ and $y$ axes, so a 1:1 correspondence would from a diagonal from bottom-left to top-right of the plot. In each case we show the results of a linear regression, with 95\% confidence intervals. Thus the values for each classifier-dataset pair as estimated after a few, then all, samples constitutes a single point marked on the plot. For each combination, the vertical distance between the actual point and the mean regression line shows the difference between the value as measured from all samples available, and the value predicted on the basis of just $n$ samples .

From  Figure~\ref{figure-real-webb100} we make the following observations:
\begin{itemize}
\item The models built from only 100 samples do not fit the data well: the plots are very scattered and the coefficient of determination is low -- in other words the linear regression shown would only account for 31-32\% of the observed variation in values for the final variables (bias, error, variance).
\item The bias terms account for the majority of the observed error.
\item Comparing the  estimates of variance after $n=100$ with the final values, the former are much higher.  This makes it apparent that the small size of the data sets is leading to considerable noise, which introduces error into the modelling process.
\item If we visualise a diagonal line through the plots for variance and total error, in each case the regression line lies below this -- so the models show the observed values with $n=100$ overestimate the final values.
\item For the bias plot the markers for all sized datasets would fall fairly evenly on either side of the diagonal. Thus the ``noise'' in the bias plot does not seem to be particularly a function of the dataset size.
\item By contrast, for the error and variance the markers for $n=n' = 1534$, which fall at the lower end of the scales, would fall around, or often above the 1:1 line, whereas those for the larger data sets would predominantly fall below the line.
\end{itemize}

This last observation is worthy of further consideration. It shows that the linear regression is a compromise. For the smaller datasets ($n+n'=1534$) whatever form the variance takes as a function of $n$, a Taylor expansion would give similar values to those observed after ($n=100$), whereas for the larger datasets the variance clearly falls away. However as the distribution of actual values for different datasets of the same size $n'$ is wide, and overlaps those for different $n'$,  it is not possible for a single regression line to capture the differences.

Turning our attention to Figure~\ref{figure-real-webb1000} we see a very different picture:
\begin{itemize}
\item   The close fit of the models built from 1000 samples to the observed data  can be confirmed both visual inspection (all points fall very close to the regression line), and statistical analysis (coefficient of determination shows that for bias and total error, the model accounts for 94\% of the observed differences).
\item The variance accounts for a smaller proportion of the total error.
\item There is no clear difference between the results for different values of $n'$.
\end{itemize}

This last observation is perhaps the least expected: if our arguments about the Taylor expansion of variance for $n=100$ hold true, they should do even more so for $n=1000$ so we might see the difference in the distribution of variance markers for different sized datasets to be even more extreme.
The fact that it is not can be explained by the hypothesis that the variance follows some inverse power-law in $n$ -- as suggested by Eq.~\ref{eq:cortes-pwerlaw}. Intuitively,  if elements of this variability are caused by the presence or absence in the training set of samples from particular regions of the data space, then both the probability of such elements not being present, and the averaged effect of their influence, fall non-linearly as $n$ increases.

However, the major point to be emphasized here is that even using a very simple model that is a linear regression from observed quantities, and does not take into account how far into the future one is trying to predict ($n'$ ), the models capture the characteristics of the observed data very closely. The results in Figure~\ref{figure-real-webb1000} thus form strong evidence to confirm our original hypothesis - that the behaviours of the bias and variance, although different are predictable.

\begin{figure}[tbp]
    \centering
\includegraphics*[width=\textwidth,height=10cm,bb=25 240 745 730]{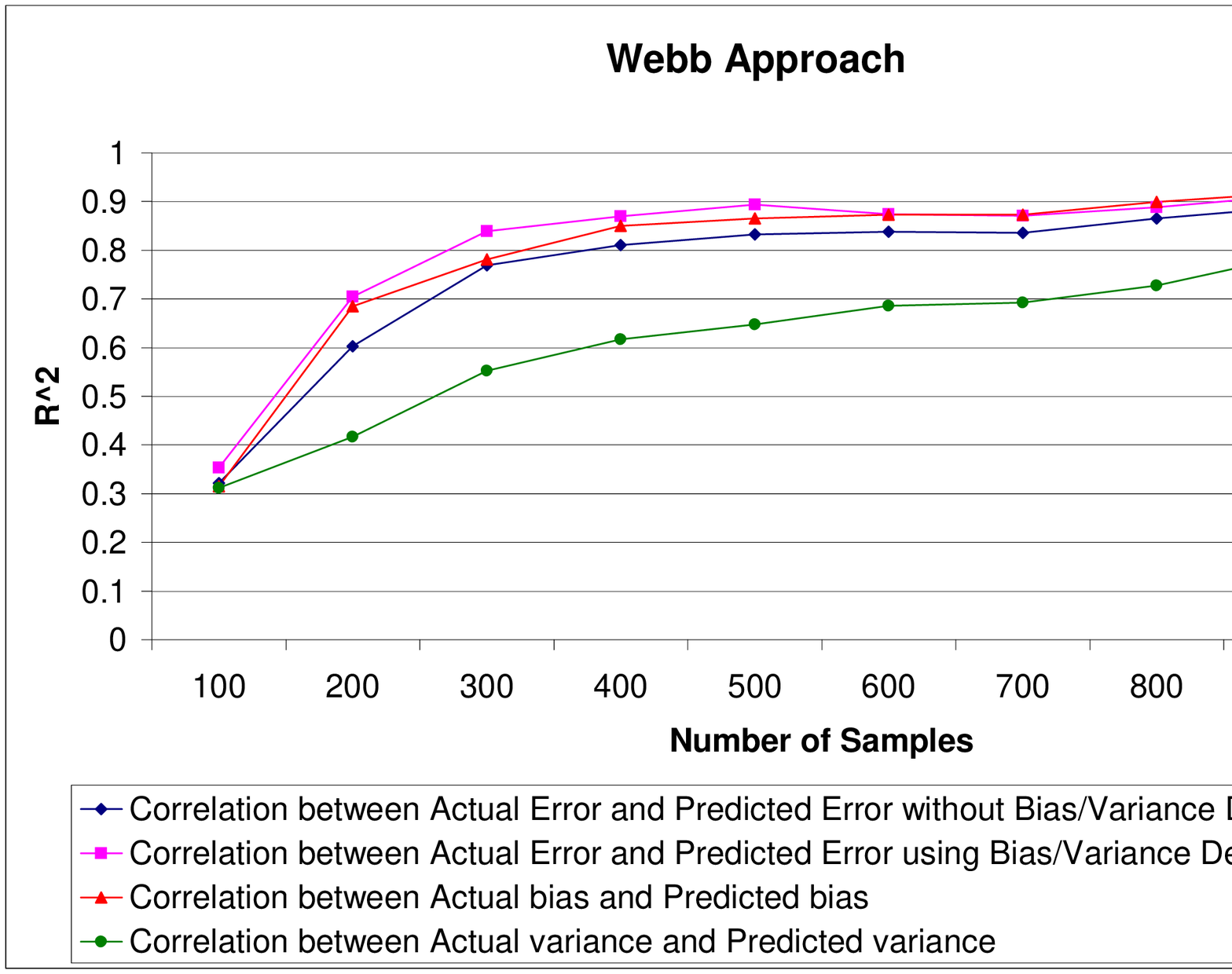}
   \caption{ Coefficient of Determination ($R^2$) between predicted and observed final error, and $n$, as a function of the number of samples used to  build the statistical models.  \label{figure_correlation}}
\end{figure}

To show how the predictive quality of the models changes as they are built from increasing numbers of samples, Figure~\ref{figure_correlation} shows  the coefficients of determination computed during the regression process as a function of $n$.
For each value of $n$, the bias, variance, and total error are estimated using SSCV as before and regression models  built relating these to the final observed values. For each data point at $n$ samples we then obtain the predicted final error in two ways: either directly using the error-error models or by calculating and then summing the predicted final bias and final variance.  It is clear that the use of separate models for bias and variance  provides better estimates of the predicted error. The plot also shows how rapidly the estimates (and correspondingly the predictive quality of the regression) stabilise in these two cases.

\section{Results: Predictive Value of Models for Unseen Data and Algorithms}\label{unseen}
We now turn our attention to examining how well the statistical models built with one group of dataset-algorithm combinations is able to predict the behavior using previously unseen datasets, algorithms, or both.

\subsection{Using the Models to Predict Error for Previously Unseen Datasets}
\begin{figure}[tbp]
\setlength{\unitlength}{0.4cm} 
\centering 
\begin{picture}(30,15) 

\put(2,13){$n$ samples}
\put(2,12){SSCV}
\put(2.5,10){\vector(2,1){5}}

\put(7.5, 10){\framebox(10,5){\begin{tabular}{l}Estimated Bias ($b_{ixn}$) \\ Estimated Variance ($b_{ixn}$)\\ Estimated Error ($e_{ixn}$)\end{tabular}}}

\put(20,12){Estimate from}
\put(22,11){Regression Models}
\put(17.5,12.5){\vector(2,-1){5}}

\put(25,5){\vector(0,-1){2}}

\put(0,5){\framebox(5,5){\begin{tabular}{l}	Classifier $i$ \\ Dataset $x$ \end{tabular}}}
\put(18,5){\framebox(14,5){\begin{tabular}{ll}Bias:& $P_b = a_{Bn}* b_{ixn} + b_{Bn}$ \\
																						 Variance:& $P_v = a_{Vn}* v_{ixn} + b_{Vn}$\\
																						 Error (1):& $P_{e1} = P_b+P_v$ \\
																						 Error (2):&$P_{e2} = a_{En}* e_{ixn} + b_{En}$ \end{tabular}}}

\put(1,2){$n + n'$ samples}
\put(1,1){ SSCV}
\put(2.5,5){\vector(2,-1){5}}
\put(7.5, 0){\framebox(10,5){\begin{tabular}{l}Estimated Bias ($b_{ixn'}$) \\ Estimated Variance ($v_{ixn'}$)\\ Estimated Error ($e_{ixn'}$)\end{tabular}}}
\put(17.5,2.5){\vector(1,0){5}}
\put(23,2.1){\textbf{compare}}
\end{picture}
   \caption{Methodology for evaluating predictive models. \label{fig:prediction}}
\end{figure}
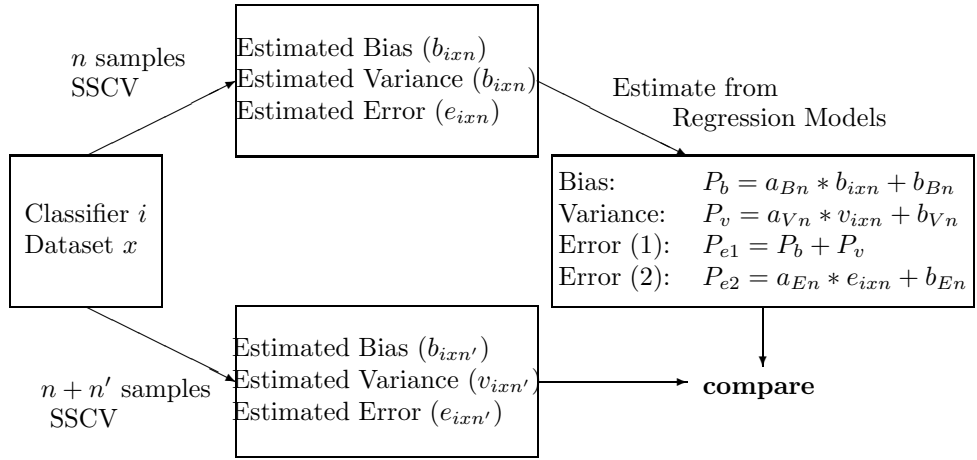

In order to evaluate the predictive capability of the models for the combination of previously used algorithms and  previously unseen data we proceeded as follows:
\begin{itemize}
  \item For each combination of the ten classifiers, the unseen data sets, and the ten sample sizes we repeat the following:
  \begin{itemize}
    \item Estimate the error and its bias and variance components using SSCV.
    \item Use these with the regression models created above to predict the  final error either directly (i.e. using error-error regression models) or via decomposition.
    \item Compare these predictions to the observed final error (Actual Error).
  \end{itemize}
  \item Perform a regression analysis between the predicted and observed values for the final error, as a function of the number of samples used to make the predictions ($n$), comparing the effects of using a single model for error or the decomposed models.
\end{itemize}
This procedure is illustrated in Figure~\ref{fig:prediction}.

\begin{figure}[tbp]
    \centering
		\includegraphics*[height=0.8\textheight]{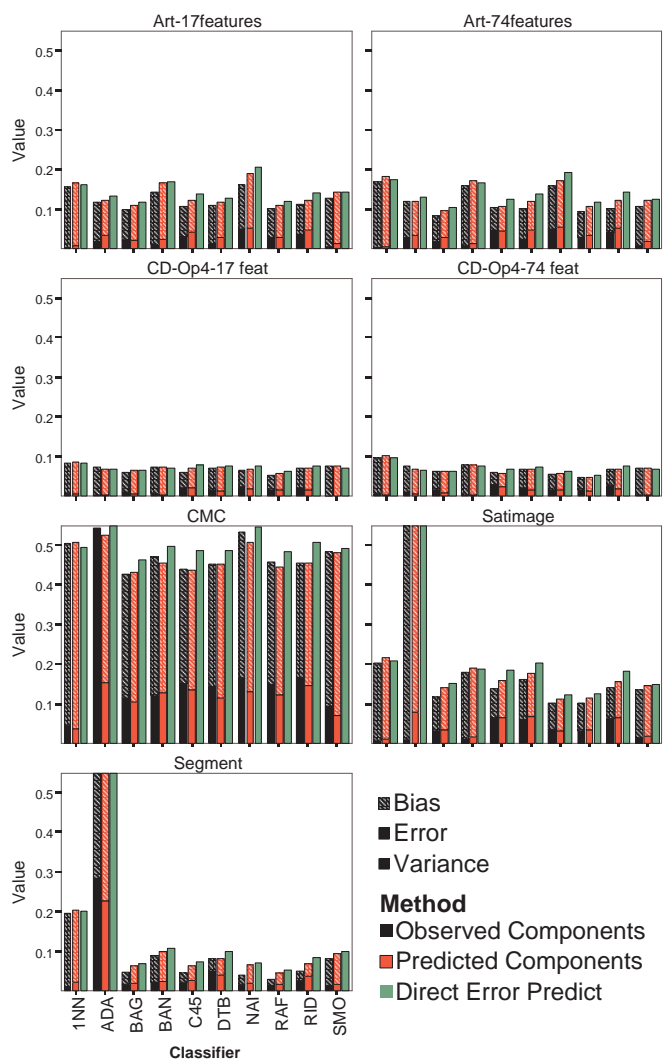}
    \caption{Comparison of error observed after $n+n'$ samples (1st, black bar)with that predicted from $n=1000$ samples using decomposed (middle, red bar) or direct (right, green bar) prediction. For observed error and decomposed predictions, stacking within bars shows bias and variance components.  \label{fig:unseendata_accuracy}}
\end{figure}

Figure~\ref{fig:unseendata_accuracy} shows the observed errors and the predicted values with and without bias-variance decomposition using the 10 different classifiers described in Section~\ref{methodology:choice} and seven previously ``unseen'' datasets. The first four are  image processing datasets Artificial~04 and CD-Operator~4 respectively with the two sizes of feature space. To investigate how well the models can extrapolate  when the initial observed accuracies lie outside the range of values ($\sim 3-29\%$) used to build the models, and when the datasets come from very different problems described by different numbers of features, we also used three  well known data sets from the UCI repository \cite{Repository}. These were  Satimage (4435 samples, 36 features), Segment (2310, 18) and CMC (1473, 9).

As can be seen, there is a close alignment between predicted and observed values and the method correctly indicates those cases (e.g.\ ADABoost-Satimage, AdaBoost-Segment) where the accuracies are low. This is a good example of providing an ``early warning mechanism'' that some remedial action might be needed. In general the predictions made via decomposition are more accurate than those made directly using the error-error regression models.  Specifically, the predictions of bias are highly accurate but there is a tendency for the variance terms to be overestimated, leading to an overestimation of the combined error.

The most noticeable errors in prediction occur on the CMC dataset which has 9 features (so is relatively ``dense'') and a total of 1437 data items. In this case there is a noticeable effect that the variance term is underestimated for most classifiers. This could be because the variance term is known to decrease with $n'$ -- as echoed by the model $P(Var) = 0.82 \cdot Var(1000)$ --  and the value of $n'$ for this dataset (473) was lower than for any of the data used to build the model. However, this effect of overestimating the variance is not apparent for the CD datasets ($n' = 543$). A more likely explanation is that the CMC dataset shows higher errors than were used to create the initial regression models, which suggests a weakness in extrapolation.  It remains for future research to examine whether  adding additional data during the model building process would create better linear regression models, or cause a need for more complex models.

\begin{figure}[tbp]
    \centering
    	\includegraphics[height=0.8\textheight]{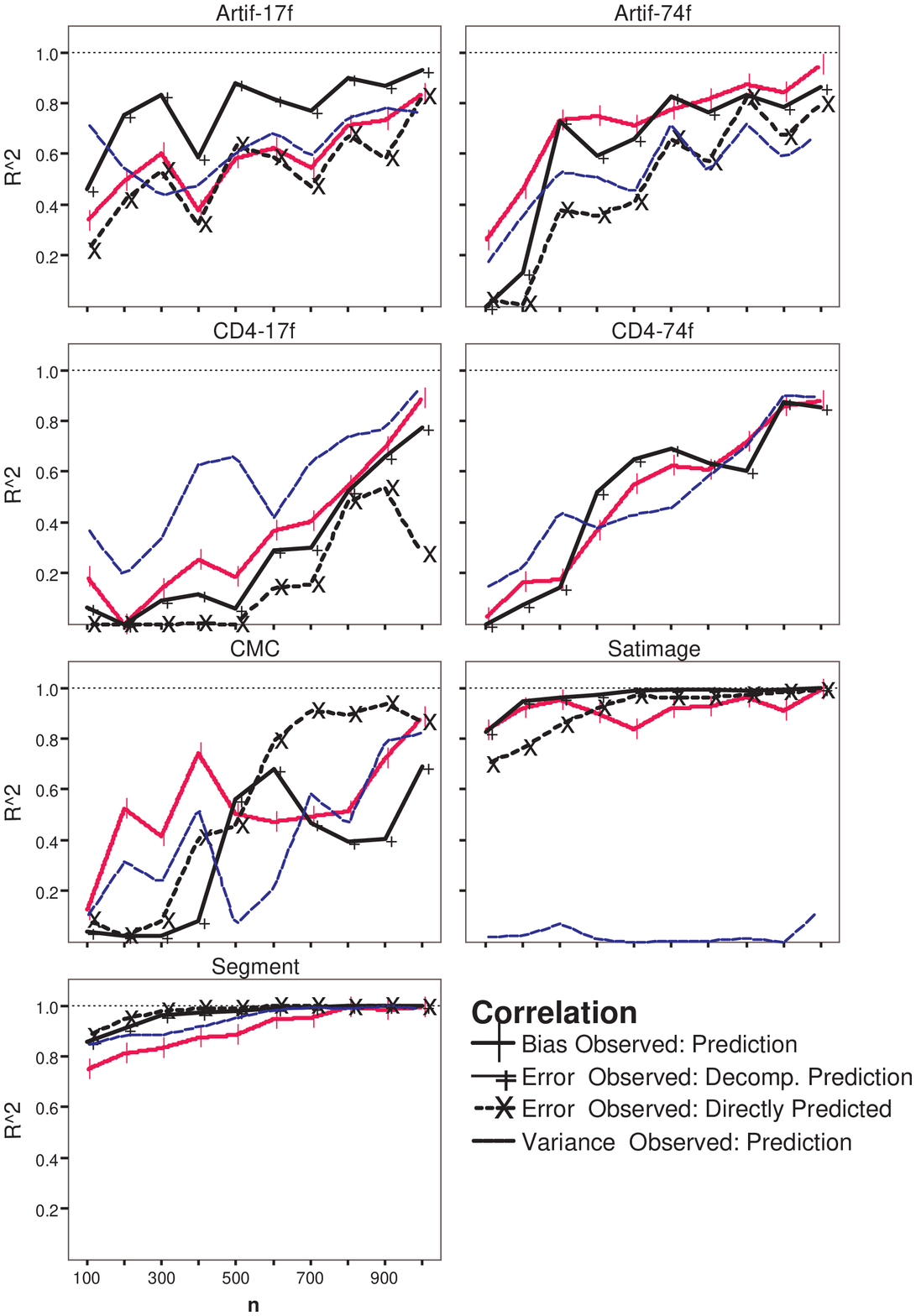}
      \caption{Relationship between sample size $n$ and the correlation coefficient between observed and predicted errors using ten ``known'' classifiers. \label{fig:unseendata_correlation}}
\end{figure}

To quantify the accuracy of the predictions, for any given dataset, we can pool the results for the ten classifiers, perform a regression analysis and then calculate the Coefficient of Determination ($R^2$) between the observed and predicted values of error, bias and variance. Figure~\ref{fig:unseendata_correlation} shows the progress of $R^2$ against $n$. Because of the small number of samples, there is a lot of ``noise'' and the curves are not monotonically non-decreasing. However, common patterns can be seen -- in each case the four values of $R^2$ rise fairly steadily with $n$, and in most cases reach values of $R^2 > 0.9$ before $n = 1000$. The exception is the variance for Satimage, which is consistently low. Notably, for the larger datasets (Artif, Segment and  Satimage) the correlations between the observed errors for different classifiers,  and those predicted via decomposition has $R^2 > 0.8$ for $n \geq 500$, and in fact $R^2> 0.9$ for $n \geq 300$. In every case except CMC, the correlation is consistently higher for decomposed error predictions than for direct predictions.
To conclude,  when using 1000 samples to make predictions of the classifier accuracy attained after using the full training set, with a set of ten diverse algorithms and seven datasets, the coefficient of determination between the predicted and observed errors is greater than 90\%.

To further evaluate the predictive performance, we treated each classifier-dataset combination as a potential item with pairs of (predicted-observed) measurements for bias, variance and total error. Since there may in principle be a huge number of possible datasets used, it is reasonable that all of these measurements may be considered samples from  an underlying normal distribution, and therefore it is appropriate to use  paired samples T-tests. These confirm that at the 95\% confidence level, the deviations for variance and directly predicted-observed error are significant. In contrast, there is a less than 0.01\%  probability that  the deviations for bias and error predicted via decomposition and are only significant.

\subsection{Using the Models to Predict Error for Previously Unseen Datasets and Algorithms}

\begin{figure}[tbp]
\centering
\includegraphics*[height=0.8\textheight]{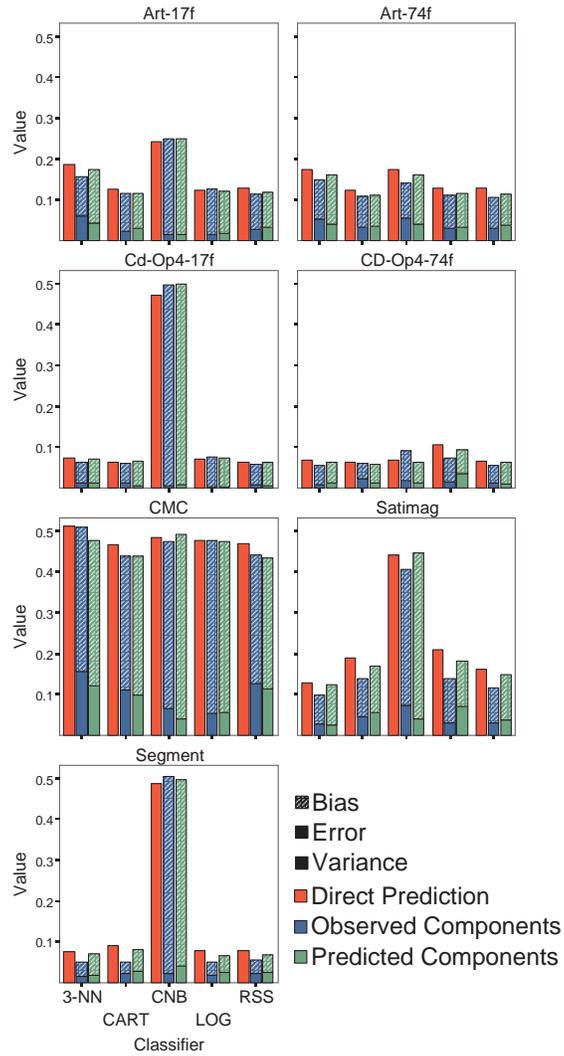}
\caption{Comparison of error observed after $n+n'$ samples (middle, blue bar) with that predicted from $n=1000$ samples using decomposed (right, green bar) or direct (left, red bar) prediction. For observed error and decomposed predictions, stacking within bars shows bias and variance components.
\label{fig:error-unseenboth}}
\end{figure}

\begin{figure}[tbp]
\centering
\includegraphics*[height=0.8\textheight]{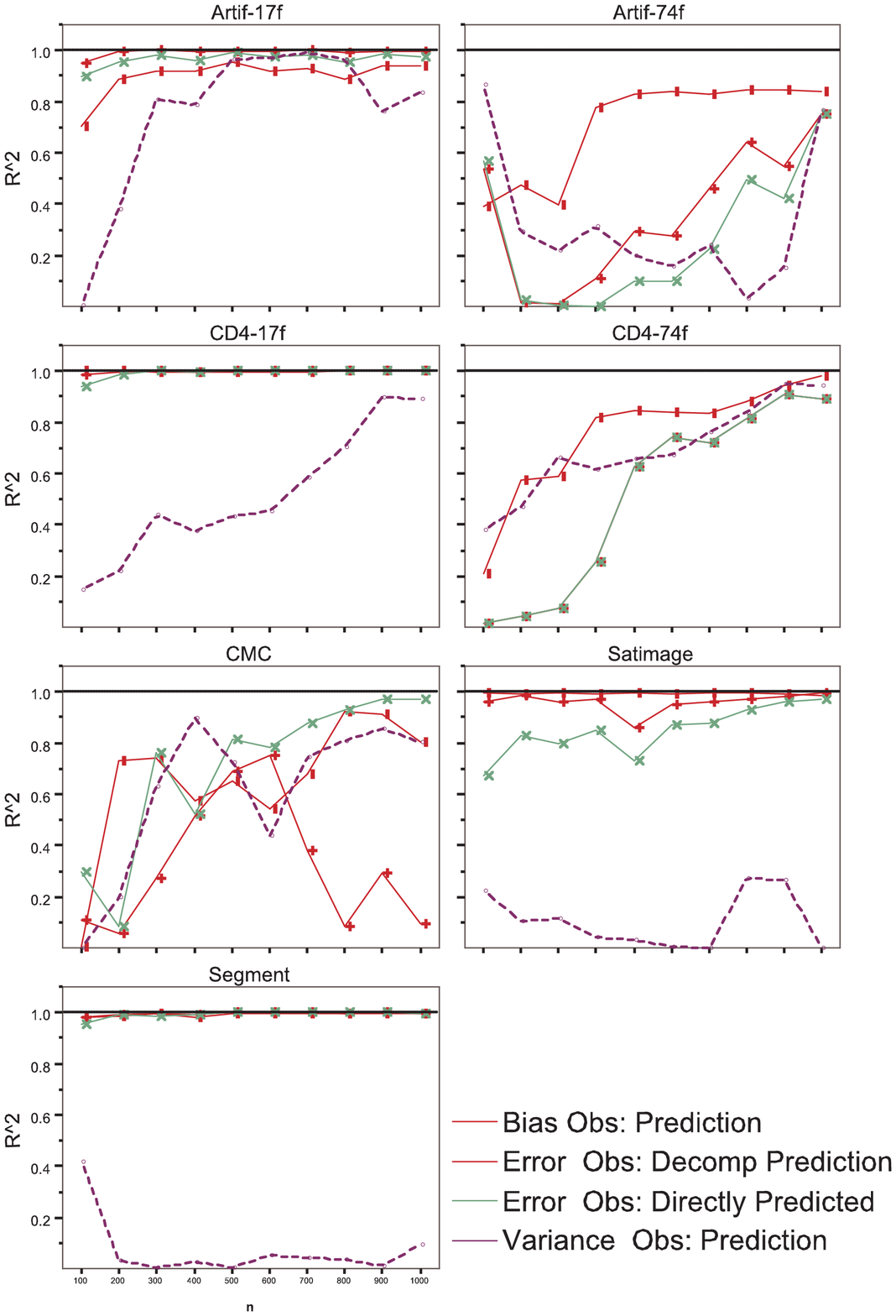}
\caption{Relationship between sample size $n$ and the Correlation Coefficient  between observed and predicted errors using  ``unknown'' data sets and algorithms.
\label{fig:unseenboth_correlation}}
\end{figure}

We next investigate how well the models can extrapolate when new algorithms are used to build classifiers, i.e.\ ones not used during training. Five algorithms are used for this analysis, namely CART (CART) \cite{LBreiman}, RandomSubSpace (RSS) \cite{Tin}, Logistic (LOG) \cite{LeCessie}, KNN (3NN), and Complement Naive Bayes (CNB) \cite{Rennie}. All these  are implemented in the WEKA library \cite{Weka}, and the default parameters in WEKA are used for each.

Figure~\ref{fig:error-unseenboth} compares the final observed  error to the predictions made with and without bias-variance decomposition for the 5 new algorithms building classifiers for the seven ``unseen'' datasets. As seen before, in almost every case the use of the separate model for bias-variance provides better estimates of the error than than the simple error model without decomposition. As before, the exception is the CMC. Nevertheless, in all cases, the predicted error is near to the actual error -- note in particular the Complement Naive Bayes (CNB) -- Satimage and Segment results.

Again we hypothesise that the \textit{relatively} poor predictive accuracy on the CMC dataset arises from the very high bias components, and inaccurate extrapolation by the regression model from its original data to these high values. Nevertheless it is worth pointing out that the decomposed approach correctly predicts the final rank order of the five classifiers.

Figure~\ref{fig:unseenboth_correlation} shows the coefficient of determination ($R^2$) between predicted and observed error as a function of the number of initial samples. Again, the small number of observations (5 -- one for each unseen classifier) for each value of $n$ cause some noise, but the correlation is high and stable for the Artif-17f, Artif-74f, Satimage and Segment data sets, rises for the Artif-74f and CD-74f sets, and is more variable so for CMC, where the variance models do not perform well.

Running the paired samples T-test as before showed that with more than 95\% confidence the differences  between predcited and observed values are not significantly different except for variance.

\begin{figure}[tbp]
\centering
\begin{tabular}{ll}
\includegraphics*[width = 0.4\textwidth,height=0.2\textheight]{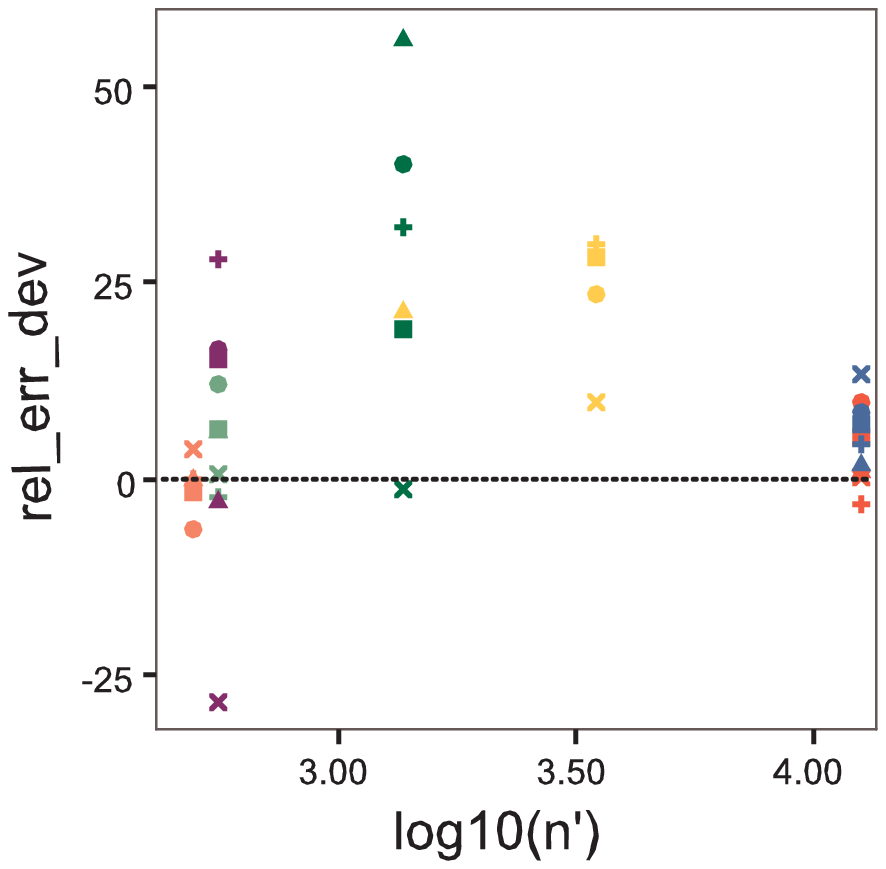} &
\includegraphics*[width = 0.4\textwidth,height=0.2\textheight]{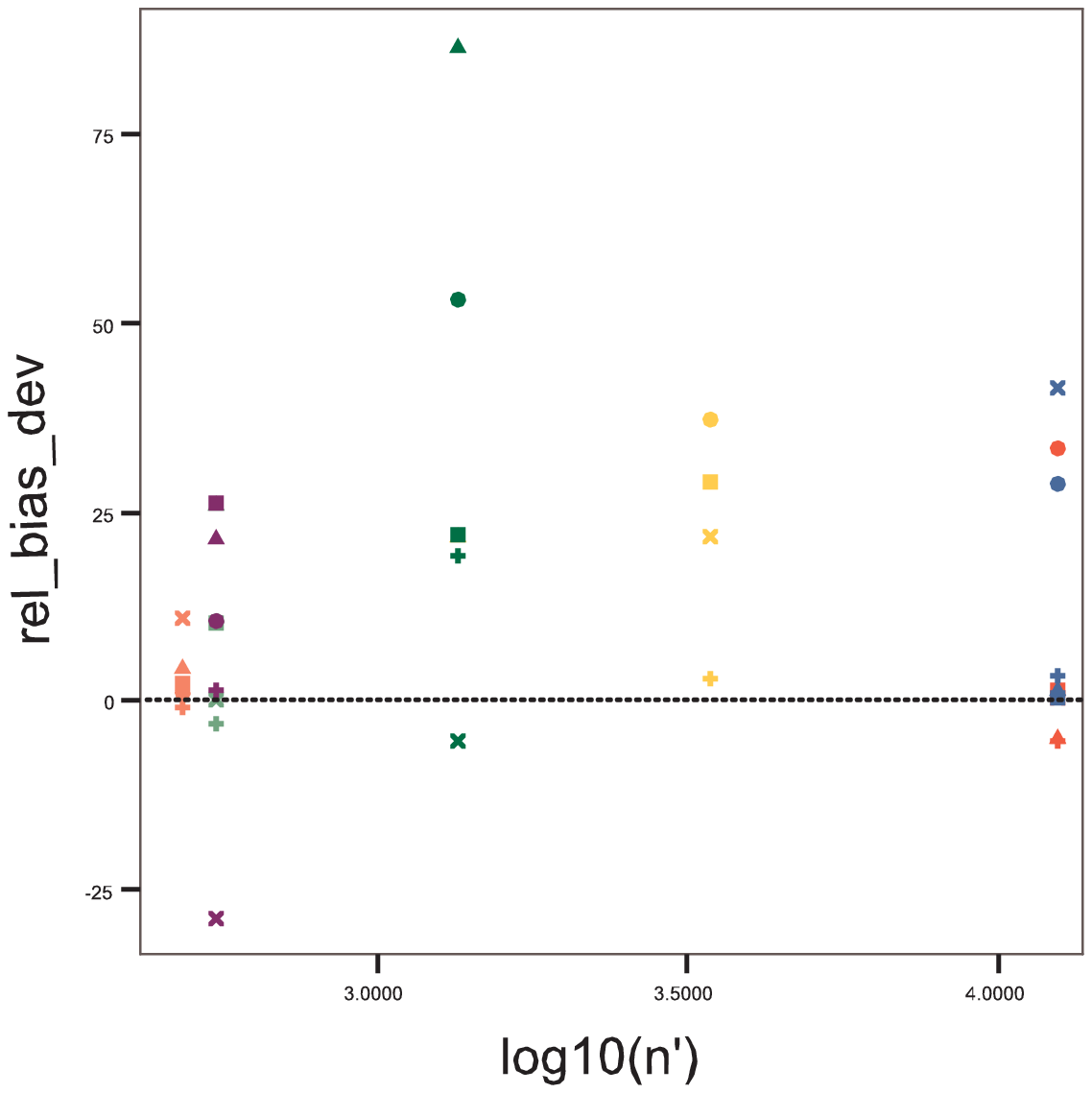} \\
\multicolumn{2}{l}{\includegraphics*[width = 0.65\textwidth,height=0.2\textheight]{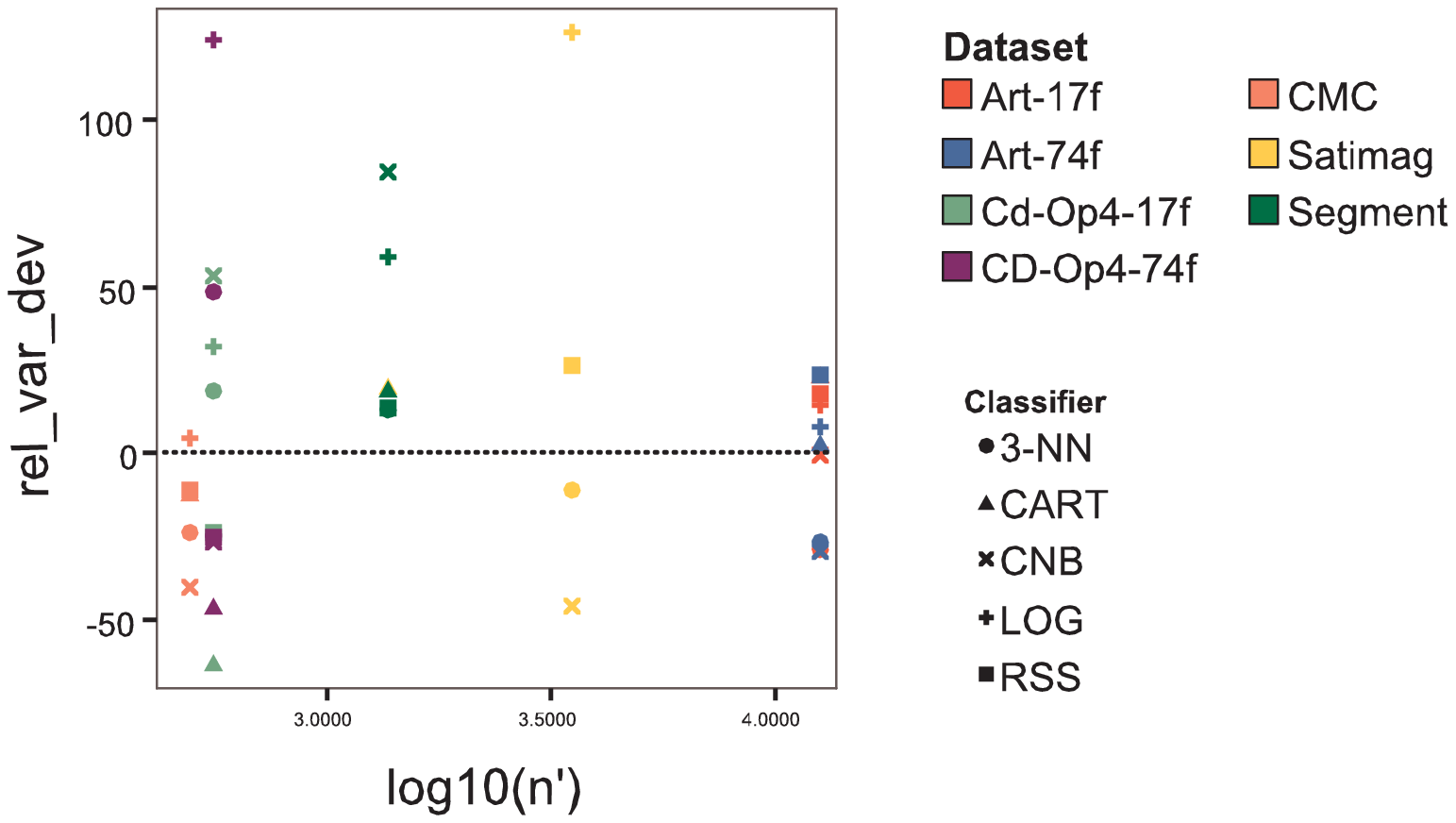}}
\end{tabular}
\caption{Relative deviations in prediction for error (top left), bias (top right) and variance as a function of $n'$. Note different scales on y-axes.
\label{fig:relative-byQ}}
\end{figure}

As a further way of analysing the data, the ``relative'' differences (i.e. $100 \cdot(prediction - observation)/observation$ were calculated and plotted for the variables error, bias and variance. Figure~\ref{fig:relative-byQ} shows plots of these values against the size of the ``extra'' data $n'$, with colours and markers to distinguish between data sets and classifiers. Note the logarithmic $x$-axis, and the different scales on the y-axes which somewhat exaggerate the deviations. Visually, there appears to be a slight trend towards overestimating bias that increases with the value of $n'$, and this causes a lesser but corresponding trend in the behaviour of the error predicted via decomposition. There is no apparent pattern for variance. However the influence of these trends are not borne out by statistical analysis -- performing a linear regression showed a near zero correlation ($R^2 < 0.01$) for each variable.

\section{Extension to Ensemble Classifiers}\label{ensembles}

The concept of decomposing error into different terms has also been used to help explain the behaviour of ensembles of algorithms.
When the algorithms concerned are performing regression tasks, decomposing the error of an ensemble into terms representing the mean bias and variance of the individual algorithms, and the covariance between them is fairly straightforward. A good recent survey of both the bias-variance-covariance and ambiguity decompositions may be found in the first few pages of \cite{Brown05}. However, just as  defining bias and variance for 0/1 loss functions was non-trivial, and there were several versions before Kohavi and Wolpert \cite{Kohavi} created their formulation in which variance is always non-negative, the extension to handle covariance in a natural way is also problematic. To the best of our knowledge there has not been a successful model decomposing 0/1 loss functions for ensembles of classifiers, so it is not immediately possible to simply extend the approach we took for single classifiers. However, in this section we present some initial findings from an approach in which we treat the entire ensemble as a single classifier. Revisiting the definitions of bias in Section~\ref{bv_decomposition}, we next develop predictors for upper limits on its attainable accuracy based on simple observations of the behaviour of the individual classifiers in the ensemble.

\subsection{Estimating Lower Bounds on the Bias for Finite Data Sets} \label{sec:ens-lower_bounds}
The analysis in Section~\ref{bv_decomposition} used a very general model predicated on the fact that the data items $x$ could be drawn from a large, potentially infinite universe of samples, corresponding to unlimited future use of the classifiers. Here we are concerned with the more limited case where  our future estimates are still drawn from a finite set of size $P+Q$.  In particular  we consider whether we can predict the values of those estimates, before completing the training process. In order to achieve this we can reformulate the models above slightly as follows.

To start with, let us assume that we have a finite set $X$ of sample data points. For consistency with above note that $|X| = P$. Because we are treating the ensemble as a single high-level entity, we need not worry about the effects of Boosting or Bagging approaches to creating ensembles by repeatedly sampling from training sets. Therefore, we assume that at our higher level training sets of size $m$  are created by sampling from $X$ uniformly without replacement. Let $D$ denote the set of training sets created in this way, and $d$ be any member of $D$, then we note that under these conditions $ P(d|f,m) = \frac{1}{|D|} = \frac{m!(P-m)!}{P!} $.

Now let $A^+, A^-, B$ partition $X$ such that  $A^+ \cup A^- \cup B = X$, where:
\begin{itemize}
\item $A^+$ is the (possibly empty) subset of data items where for all training sets a classifier trained on that set correctly predicts  the class of item $x$.
      \begin{eqnarray*}\forall x \in A^+, d,d' \in D, y \in Y &&   Y_H(y|x,d) = Y_H(y|x,d') = Y_F(y|x)\end{eqnarray*}
\item $A^-$ is the (possibly empty) subset of data items where for all training sets a classifier trained from that set incorrectly predicts  the class of item $x$.
      \begin{eqnarray*}\forall x \in A^-, d,d' \in D, y \in Y &&  Y_H(y|x,d) = Y_H(y|x,d') \neq Y_F(y|x)\end{eqnarray*}
\item $B$ is the (possibly empty) set of data items where $Y_H(y|x,d)$, the hypothesis describing the predicted class of item $x$ depends on the choice of training sets $d$.
        \begin{eqnarray*}\forall x \in B &&  \exists d,d' \in D \bullet Y_H(y|x,d) \neq Y_H(y|x,d')\end{eqnarray*}
\end{itemize}

So now lets look at what this means in terms of our estimates of the bias of the classifier. This will of course depend on the methods used for the estimates. Following well-established previous research, we will assume that each item in the data set is predicted exactly $k$ times. This is true with $k=1$ for $N$-fold cross-validation, and for $k>1$ for the Webb and Collione approach, in general although interestingly not for the Kohavi approach \cite{Kohavi}. This means that when we sum over the data items $x$ in the counterpart of Eq.~\ref{kohaviandwolpert} each term occurs with equal probability.

Note that $bias_x$ as stated above is composed of terms which themselves depend on the choice of training sets, and that we are assuming a fixed set of data points and a fixed size training sets. We therefore  refine the definition of bias to  take these into account, and average  over all possible training sets.

\begin{equation}
bias^2  =   \frac{1}{2}\sum_{x \in X}P(x)\sum_{d \in D}P(d|F,n)\sum_{y \in Y}[P(F(x)=y)-P(H_{Cd}(x)=y)]^2
\end{equation}
If we assume we are sampling iid then $P(x) = 1/|X|$ and $P(d|F,n) = 1/|D|$. We now turn our attention to the case where each data item $x \in X$ is unambiguously associated with one of two possible class labels $y \in Y$, and we will further constrain our ensemble to output crisp decisions so that $P(H_{Cd}(x)=y) \in \{0,1\}$. Partitioning the data set $X$ as above, we note that we make use of the following conditions when performing the summation. First, the set $A^+$ does not contribute to the bias since the predicted class for this subset of items is always correct.
Second, $\forall x \in X, C, d \in D , \exists y_1,y_2 \in Y , y_1 \neq y_2 : F(x)=y_1 \wedge H_{CD}(x) = y_2$. This means that within the  partition $A^-$ for each combination of $x$ and $d$, there are exactly two values of $y$ which both contribute +1 to the summation. This yields:
\begin{eqnarray}\label{eq:biasdef1}
bias^2       & = &  \frac{1}{2} \sum_{x \in A^-}P(x)\sum_{d \in D}p(d,F,n) \cdot 2 \nonumber  \\
             & +  &   \frac{1}{2} \sum_{x \in B}P(x)\sum_{d \in D}P(d|F,n)\sum_{y \in Y}[P(F(x)=y)-P(H_{Cd}(x)=y)]^2 \\
             & = & \frac{|A^-|}{|X|} + \frac{1}{2}\frac{|B|}{|X|} \cdot \frac{n!(|X|-n)!}{|X|!}\sum_{x,d}\sum_{y \in Y}[P(F(x)=y)-P(H_{Cd}(x)=y)]^2
\end{eqnarray}
The last term will take a value between  0 and $|B|/|X|$ since for each value of $y$ the difference will be 0 for some training sets and 1 for others which the gives bounds:
\begin{equation}
|A^-|/|X| < bias^2 < (|A^-| + |B|)/|X|\label{eq:biasbounds}
\end{equation}

This reformulation makes it explicit that considering the proportion of samples which the ensemble always misclassifies  will yield a strict underestimate of the bias provided that there exist any items for which the prediction made  is dependent on the training set.
Furthermore, since according to Eq.~\ref{kohaviandwolpert} the variance term is always non-negative, we can say that the quantity $|A^-|/|X|$ constitutes a strict lower bound on the error rate of a classifier -- or an ensemble treated as a single entity.

\subsection{Experimental Approach} \label{sec:ens-exper_approach}

Previous sections illustrated the successful use of regression models built from a variety of dataset-classifier combinations to predict the error rates that could be attained after future training. However, decomposing the error into different components  is not straightforward for ensembles of classifiers \cite{Brown05}. Moreover, this would require running $N$-fold cross validation a number of times to get accurate estimates of bias and variance components for each combination of dataset, algorithm, and $n$. This becomes computationally expensive when extended to a heterogeneous ensemble, particularly if the ensemble is itself trainable.

For this section we use a slightly different approach. Previously we pooled the results from many experiments to build regression models relating observations of bias and variance after different values of $n$ training data to the same variables of $n+n'$ items. Here we treat each data set independently, and built regression models to characterise the ensemble's learning curve  as a function of $n$. As noted above, there is theoretical \cite{vapnik00} as well as empirical \cite{cortes94,mukherjee03,Duda} evidence that these learning curves have a power-law dependency on the number of training samples, i.e.\ they are of the form
\begin{equation}\label{eq:regress}
    error_{ensemble} = a \cdot n^b + c \enspace ,
\end{equation}
where $a$ is the learning rate, $b$ the decay rate and $c$ the Bayes error (the minimum achievable error or, in the error-decomposition framework, the ``noise'').

In our experiments, the bound on the ensemble's error derived in Section~\ref{sec:ens-lower_bounds}, $|A^-|/|X|$, was used as an estimate of the minimum achievable error. When faced with a new dataset-ensemble combination, we make observations of $|A^-|$ and the ensemble error at regular intervals, and then feed these into the power-law regression model in order to fine-tune the parameters of the model so that it fits the new data and predicts the future development of the ensemble error, as will be detailed in Section~\ref{sec:results-learning_curves}. Before elaborating on these results, in Section~\ref{sec:results-oracle} we will provide an analysis of the stability of the estimation of the lower bound on the error by using $|A^-|/|X|$.

\subsection{Analysis of the Stability of Estimators of Lower Bounds on Error} \label{sec:results-oracle}

For the experiments performed here 22 Machine Vision datasets from the DynaVis project were used (2 different feature spaces -- 17 and 74 features -- for each of 5 CD-Print, 5 Artificial, and the Egg image sets). The CART \cite{LBreiman} and C4.5 \cite{Quinlan93} decision trees, the Naive Bayes \cite{Duda}, Nearest Neighbour \cite{Cover} and eVQ \cite{lughofer08} classifiers were used as base classifiers, the decisions of which were combined using the Discounted Dempster-Shafer ensemble training method \cite{sannen07}. For each data set, each classifier, and each value of $n \in \{100,120,\ldots,1000\}$ samples, $N$-fold cross validation was repeated $l$ times to make $l$ predictions of the class of each item in the training set. From this data we calculated the values of $|A^-|/|X|$ as a function of $n$ for each data set (i.e.\ 22 values for each value of $n$). For clarity we denote the values $|A^-|/|X|$ hereafter as $Or_{n}$.

\begin{figure}[tbp]
	\centering
	\includegraphics[scale=0.6]{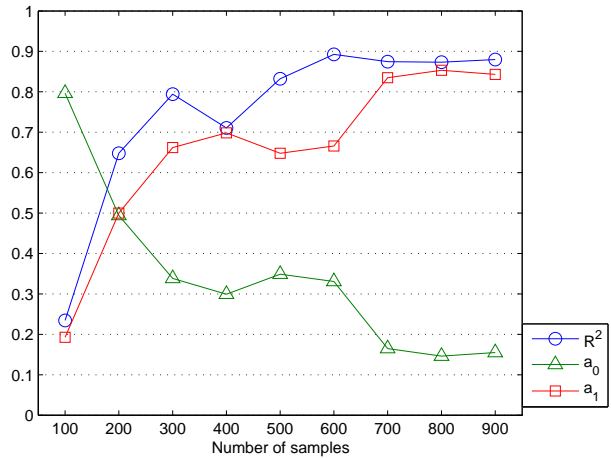}
    \caption{The linear regression components for the correlation of $Or_{n}$ vs.\ $Or_{final}$ together with the coefficient of determination $R^2$ as a function of $n$. 	\label{fig:oracle_progression}}
\end{figure}

In order to examine the stability of the predicted bounds as $n$ increased, we plotted $Or_{n}$ against $Or_{final}$
and used linear regression as before to fit a model of the form $Or_{final} = a_1 \cdot Or_{n} + a_0$, and to estimate the quality of the model via $R^2$. Figure~\ref{fig:oracle_progression} shows the progression of the coefficients $a_0$ and $a_1$ and the corresponding values of $R^2$ as a function of $n$.

As can be seen in Figure~\ref{fig:oracle_progression}, the models generated as $n$ increases do produce predictions which correlate well to the observed values after further training. However, as can be seen by the progression of the coefficients, the nature of the regression models changes. For low values of $n$ the models predict a high constant value for $Or_{final}$ with a low component related to the observed value of $Or_{n}$ -- essentially the system has not seen enough ``difficult'' samples. Since the major component of the predicted value of $Or_{final}$ is fixed for $n=100$, the correlation is fairly low. As $n$ increases and a more representative sample of the data is seen, the situation changes. Thus for training set sizes $n \geq 700$ the predicted value is dominated by the observed value ($a_1 \approx 0.85$) with only a low constant component ($a_0 \approx 0.15$). For these training set sizes $R^2$ increases to approximately 0.9.

\subsection{Empirical Results for Predicting Lower Bounds and Total Errors} \label{sec:results-learning_curves}
The values $Or_{n}$ for different $n$ can be used for predicting not just a lower bound on, but also an estimate of the error of a trained ensemble. The following procedure can be used:
\begin{enumerate}
\item $Or_{n}$ is measured for different $n$ and a constant regression is performed for these values, i.e.\ we obtain the constant $OR$ which minimises the Mean Square Error with the values of $Or_{n}$ across different values of $n$. This value forms our estimate of the lower bound on the achievable error.
\item The errors the ensemble makes are also recorded for different $n$.
\item A power-law regression is performed for the ensemble errors, asymptotically approaching the estimated constant $OR$ as calculated in step 1:
\begin{equation}\label{eq:regress2}
    error_{ensemble} = a \cdot n^b + OR \enspace ,
\end{equation}
where $a$ and $b$ are the regression parameters which are optimised in the regression procedure.
\item An analogous procedure is used to model the standard deviation of the observed values.
\item From the power-law regression model we can estimate the error of the ensemble after $n+n'$ samples are presented, and also some estimates of how the variation changes.
\end{enumerate}

\begin{figure}[tbp]
    \centering
    \subfigure[Results for CD-Operator~4]
    {
        \includegraphics[height=0.35\textheight]{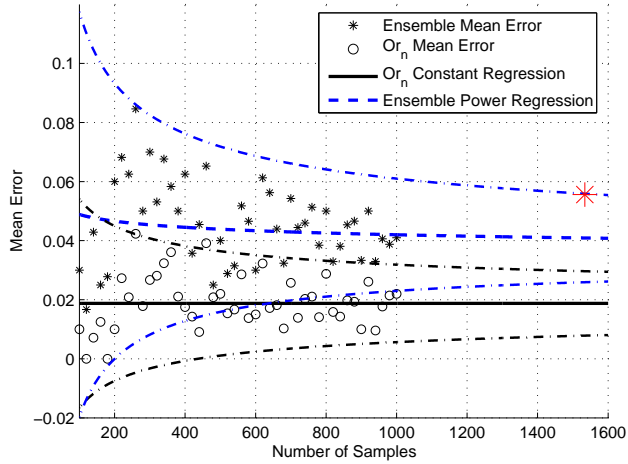}
        \label{fig:oracle_example_sony4}
    }\\
    \subfigure[Results for Artificial~04]
    {
        \includegraphics[height=0.35\textheight]{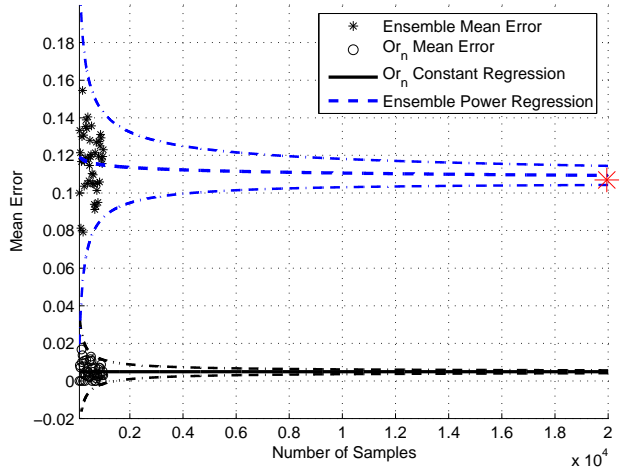}
        \label{fig:oracle_example_artif08}
    }
    \caption{Prediction of the errors of a (trainable) ensemble for $n+n'$ samples using a regression model which is built using $n$ training samples. The mean values of $Or_{n}$ and the mean errors of the ensemble are shown, together with their regression models (including standard errors). The error when training on the entire data set is depicted by the large $\ast$.}
    \label{fig:oracle_example}
\end{figure}

The results of this procedure are illustrated in Figure~\ref{fig:oracle_example}. The five base classifiers listed above are combined using the Discounted Dempster-Shafer combination ensemble \cite{sannen07}. $Or_{n}$ was measured for $n=\{100,120,\ldots,1000\}$ samples. A constant regression was performed to model $Or_{n}$ with a constant value and the obtained value is then used as an asymptote when modelling the ensemble errors. The errors the ensemble makes are again recorded for $n=\{100,120,\ldots,1000\}$ samples and a (robust) regression model is built according to Equation~\ref{eq:regress}. The results of this procedure are illustrated in Figure~\ref{fig:oracle_example_sony4} for CD-Operator~4 and in Figure~\ref{fig:oracle_example_artif08} for Artificial~04. The values $Or_{n}$ and the errors of the ensemble are shown for different $n$, as well as the regression models that are built for them, together with the estimated standard errors. Also the final error after evaluating the performance of the ensemble when it is trained on the entire data set is indicated, to show how accurately the errors are predicted for the ensemble when it would be trained using a larger number of training samples ($n+n'$).

First, in both cases the results show that the model of $OR$ does as expected form a lower bound on the error.
As can be seen from Figure~\ref{fig:oracle_example_artif08}, the use of the secondary robust regression method to predict the mean and standard deviation of the observed ensemble error (top set of curves) for the artificial data set, extrapolates well and the final observed error (large asterisk at $n+n' \approx 20000$) falls inside these values. For the much smaller CD print data set the figure is less clear, and the estimated standard errors on the predicted asymptote $Or_{n}$ (bottom set of curves) overlap those of the robust regression prediction. Nevertheless, again the observed final ensemble error lies within one standard deviation of the value predicted by the robust regression procedure.

\section{Conclusion}\label{conclusions}
In this paper, we have investigated techniques for making early predictions of the error rate achievable after further interactions. We have provided several example scenarios where the ability to do this would be of great value in practical data mining applications. Our approach is based on our observations that although the different components of the error progress in different ways as the number of training samples is increased,  the behaviour displayed by each component appeared to be qualitatively similar across different combinations of dataset and classification algorithm.
To investigate this finding, we have created a large set of results for many different combinations of dataset, algorithm, and training set size ($n$) and applied statistical techniques to examine the relationship between the values observed after partial training (with $n$ samples) and those after full training. 

Perhaps surprisingly, the results showed that in fact a simple linear model provided a highly accurately predictor for the subsequent behaviour of different components. Results confirmed our hypotheses that these could be combined to produce highly accurate predictions of the total observed error. These findings are validated using a range of datasets and algorithms which were not used during the creation of our statistical models. We have also examined the extent to which our models can reliably extrapolate  when new observations have values way outside the ranges of data used in our models. Results  such as the accurate predictions of poor performance for the CNB and AdaBoost algorithms on the Satimage and Segment datasets) show that the bias models extrapolate extremely well. However, the final predictions are slightly less accurate if the nature of the data is such that a high variance component is observed -- e.g.\ the CMC data of the UCI database. This suggests that a more complicated model, where the predicted value depends on the final number of samples available, may be necessary for variance.

As there is no bias-variance(-covariance) decomposition available for 0/1 loss functions for ensembles of classifiers, it is not straightforward to apply the methodology used to accurately predict the performance of classifiers after further training to ensembles of classifiers. We have shown how a reformulation of the bias component can provide an estimate of the lower bound on the achievable error which may be more easily computed. This is especially important when the cost of training is high -- for example with trainable ensembles of classifiers. This bound is used as an asymptote in a power-law regression model to accurately predict the progression of the ensemble's error, independently for each data set.

For future work, we will focus in two directions. First we will combine previous theoretical findings  and the successful results from the two different approaches here. Taken together they   suggest that for even more accurate predictions,  it is worth combining the linear model for bias with an inverse power law  model for variance using both the  current estimates and period over which to predict ($n'$) as factors. This can be expected to  prove particularly useful for classifiers where variance forms a major part of the observed error. Second, the work presented in this paper used  Kohavi and Wolpert's definition of bias and variance, and we will investigate whether using other definitions of bias and variance  further improve the predicted accuracy.

\section*{Acknowledgements}
This work was supported by the European Commission (project Contract No.
STRP016429, acronym DynaVis). This publication reflects only the authors'
views.

\end{document}